\documentclass[10pt,twocolumn,letterpaper]{article}

\usepackage[pagenumbers]{cvpr} 

\usepackage{floatrow}
\usepackage{comment}
\usepackage{bbding}
\usepackage{tabularx}
\usepackage{colortbl}
\usepackage{booktabs} 
\usepackage{adjustbox} 
\usepackage{multirow}
\usepackage{xspace}
\usepackage[acronym]{glossaries}
\usepackage{float}
\usepackage{wrapfig}
\usepackage{algorithm}
\usepackage{algpseudocode}
\usepackage{gensymb}
\usepackage{mwe}
\usepackage{subcaption}
\usepackage{soul} 
\usepackage{bm}
\usepackage{amssymb} 
\definecolor{codegreen}{rgb}{0,0.6,0}
\definecolor{codegray}{rgb}{0.5,0.5,0.5}
\definecolor{codepurple}{rgb}{0.58,0,0.82}
\definecolor{backcolour}{rgb}{0.95,0.95,0.92}
\definecolor{mediumtealblue}{rgb}{0.0, 0.33, 0.71}
\definecolor{darkpastelgreen}{rgb}{0.01, 0.75, 0.24}
\definecolor{azure}{rgb}{0.0, 0.5, 1.0}
\usepackage{microtype}

\usepackage{silence}
\usepackage[normalem]{ulem}
\newcommand{\inlinesection}[1]{\noindent{\textbf{#1}}}

\renewcommand{\paragraph}[1]{\vspace{.3em}\noindent\textbf{#1.}}

\newcommand{\Plucker}{Pl\"ucker\xspace}
\newcommand{\methodname}{AC3D\xspace}
\newcommand{\methodfullname}{Advanced 3D Camera Control}
\newcommand{\vdit}{VDiT\xspace}
\newcommand{\vditwithref}{\hyperref[sec:method:base-video-dit]{\vdit}\xspace}
\newcommand{\baselinename}{VDiT-CC\xspace}
\newcommand{\baselinewithref}{{\hyperref[sec:method:baseline-cc]{\baselinename\xspace}}}
\newcommand{\paramcount}{11.5B}
\newcommand{\ditxs}{\texttt{DiT-XS}}

\newcommand{\camditxs}{\texttt{camera DiT-XS}}

\newcommand{\linprobtrainspeedup}{15}
\newcommand{\retenkfull}{RealEstate10K\xspace}
\newcommand{\retenk}{RE10K\xspace}
\newcommand{\msrvtt}{MSR-VTT}

\newcommand{\video}{\bm{x}}

\newcommand{\textcond}{\bm{t}}

\newcommand{\R}{\mathbb{R}}

\newcommand{\cameracond}{\bm{c}}
\newcommand{\camextr}{\bm{C}}
\newcommand{\camintr}{\bm{K}}

\newcommand{\config}[1]{{\mdseries\textsc{Config~\small\uppercase{#1}}}}

\newcommand{\apprx}[1]{${\approx}{#1}$} 

\newcommand{\cellbest}{\cellcolor{azure!35}}
\newcommand{\cellsecond}{\cellcolor{azure!10}}
\newcommand{\ours}{\textcolor{azure}{(ours)}}
\newcommand{\abl}[1]{\textit{#1}}

\definecolor{cvprblue}{rgb}{0.21,0.49,0.74}
\usepackage[pagebackref,breaklinks,colorlinks,allcolors=cvprblue]{hyperref}

\def\paperID{x} 
\def\confName{CVPR}
\def\confYear{2025}

\title{AC3D: Analyzing and Improving 3D Camera Control \\ in Video Diffusion Transformers}

\author{
  Sherwin Bahmani$^{*1,2,3}$
  \space\space
  Ivan Skorokhodov$^{*3}$
\space\space
  Guocheng Qian$^{3}$
  \space\space
  Aliaksandr Siarohin$^{3}$\\
  \space\space
  Willi Menapace$^{3}$
  \space\space
  Andrea Tagliasacchi$^{1,4}$
  \space\space
  David B. Lindell$^{1,2}$
  \space\space
  Sergey Tulyakov$^{3}$
  \\
  \space\space
  \small{\textnormal{$^{1}$University of Toronto\space\space$^{2}$Vector Institute\space\space$^{3}$Snap Inc.\space\space$^{4}$SFU}}\\
    \small{*equal contribution}
}

\begin{document}

\twocolumn[{
\maketitle
\vspace{-3em}
\begin{center}
    \url{https://snap-research.github.io/ac3d}
\end{center}
\begin{center}

\includegraphics[width=\linewidth]{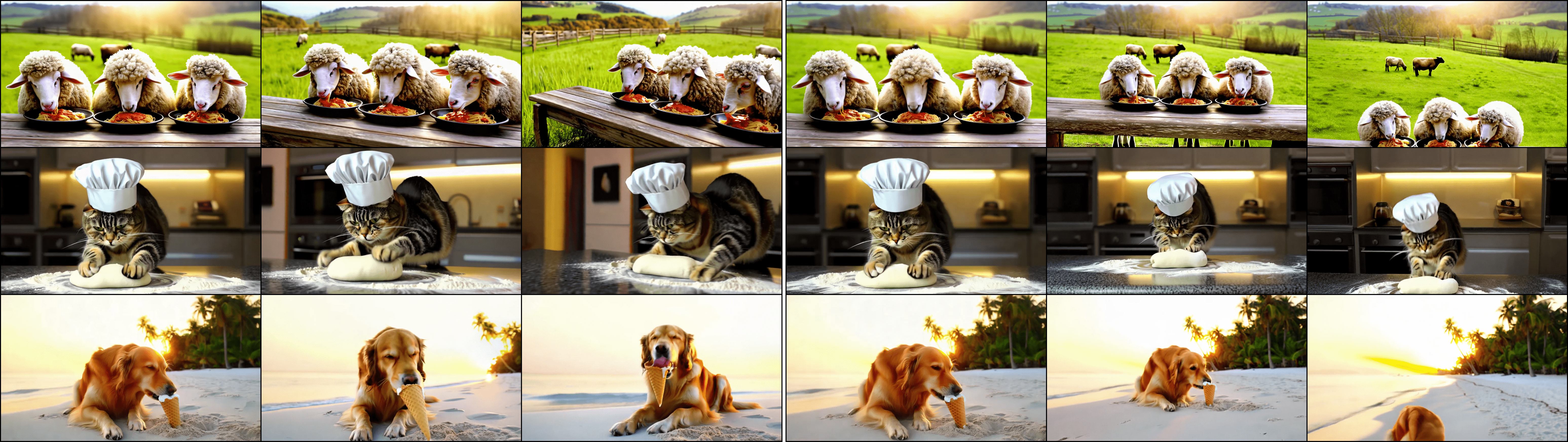}
\end{center}
\vspace{-1.5em}
\begin{center}
\begin{minipage}[t]{.195\linewidth}
\centering
\includegraphics[width=\linewidth]{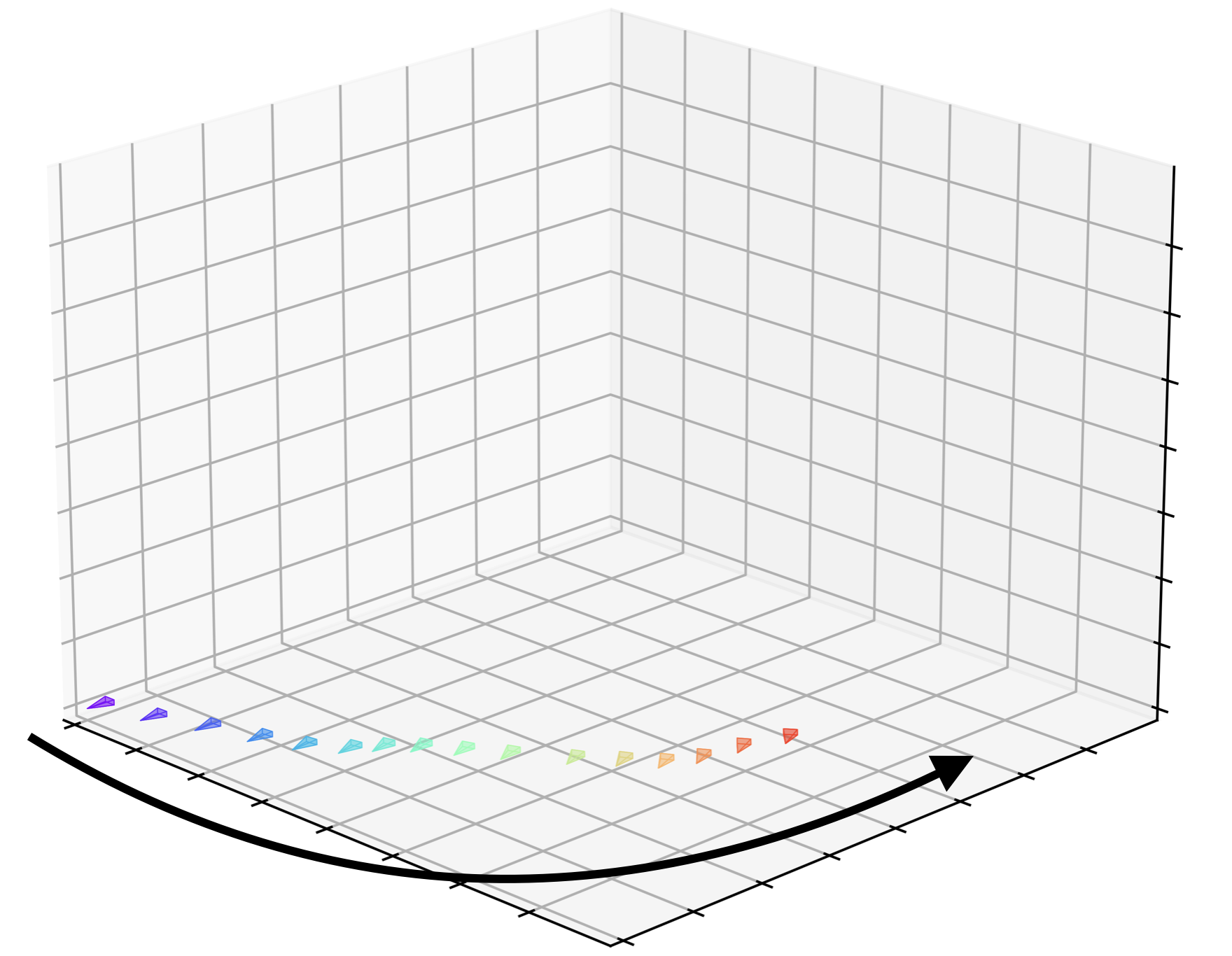}
\end{minipage}
\hfill
\begin{minipage}[t]{.58\linewidth}
\vspace{-7.5em}
\captionof{figure}{\textbf{Camera-controlled video generation.} Our method enables precise camera controllability in pre-trained video diffusion transformers, allowing joint conditioning of text and camera sequences. We synthesize the same scene with two different camera trajectories as input.
The inset images visualize the cameras for the videos in the corresponding columns. The left camera sequence consists of a rotation to the right, while the right camera visualizes a zoom-out and up trajectory.
}
\end{minipage}
\hfill
\begin{minipage}[t]{.195\linewidth}
\includegraphics[width=\linewidth,]{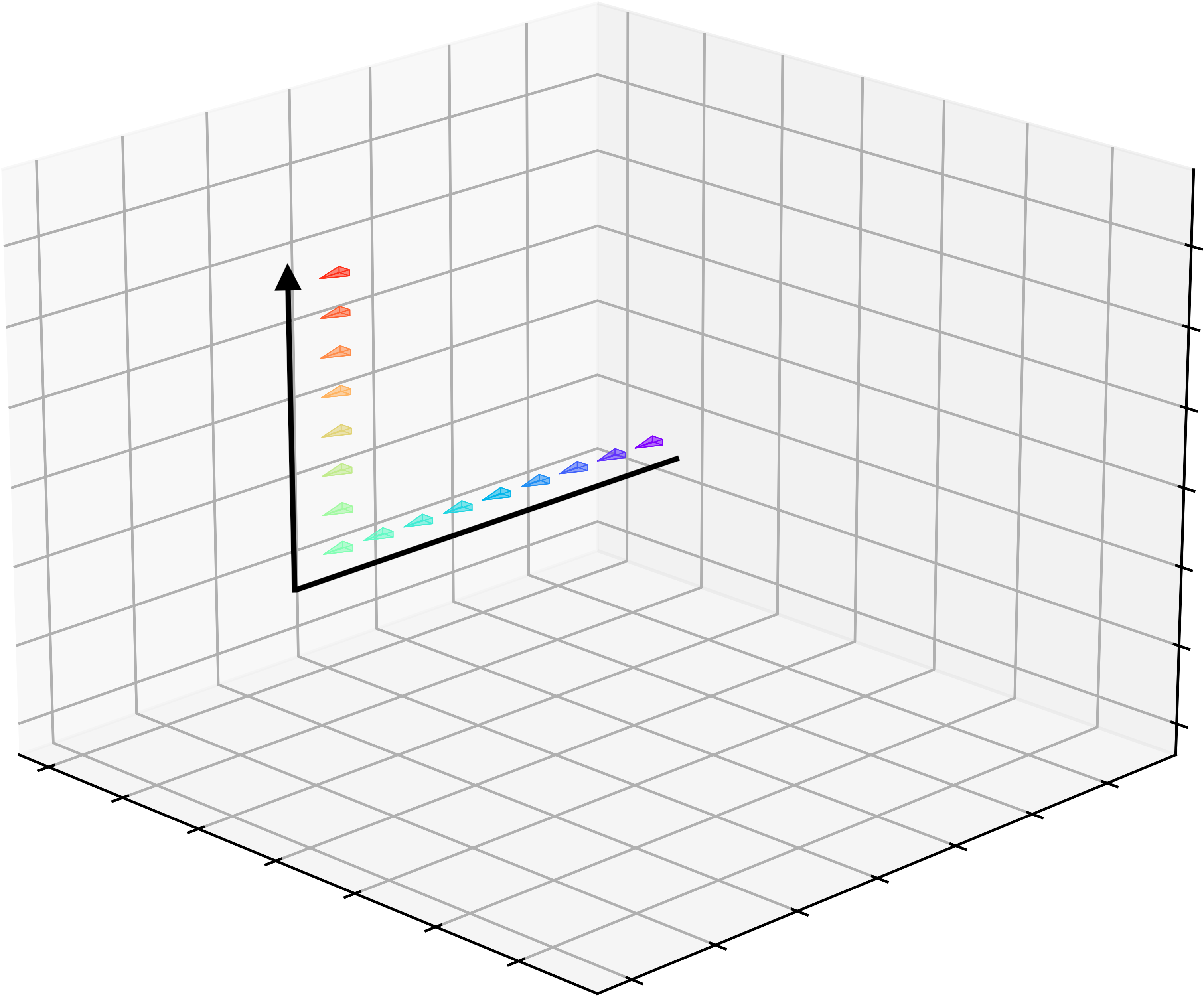}
\end{minipage} 
\label{fig:teaser}
\end{center}

}]
\begin{abstract}
Numerous works have recently integrated 3D camera control into foundational text-to-video models, but the resulting camera control is often imprecise, and video generation quality suffers.
In this work, we analyze camera motion from a first principles perspective, uncovering insights that enable precise 3D camera manipulation without compromising synthesis quality.
First, we determine that motion induced by camera movements in videos is low-frequency in nature.
This motivates us to adjust train and test pose conditioning schedules, accelerating training convergence while improving visual and motion quality.
Then, by probing the representations of an unconditional video diffusion transformer, we observe that they implicitly perform camera pose estimation under the hood, and only a sub-portion of their layers contain the camera information.
This suggested us to limit the injection of camera conditioning to a subset of the architecture to prevent interference with other video features, leading to a $4\times$ reduction of training parameters, improved training speed, and 10\% higher visual quality.
Finally, we complement the typical dataset for camera control learning with a curated dataset of $20$K diverse, dynamic videos with stationary cameras.
This helps the model distinguish between camera and scene motion and improves the dynamics of generated pose-conditioned videos.
We compound these findings to design the \methodfullname~(\methodname) architecture, the new state-of-the-art model for generative video modeling with camera control.
\end{abstract}

\section{Introduction}
\label{sec:intro}

Foundational video diffusion models (VDMs) trained on internet-scale data, acquire abundant knowledge about the physical world~\cite{Sora}.
They not only learn appearance and plausible 2D dynamics, but they also have abundant understanding of 3D structure~\cite{blattmann2023stable}.
However, most of this knowledge is stored \textit{implicitly} within the model, as these models do not expose fine-grained control mechanisms, such as camera motion control.
\quad
We recently witnessed a surge of works that bring 3D camera control into foundational video models~\cite{MotionCtrl, CameraCtrl, DirectAVideo}, but the control they provide is not very precise, and the synthesis quality is often compromised~\cite{VD3D}.
We analyze camera motion control in video diffusion models from first principles, and develop several findings that allow us to incorporate \textit{precise} 3D camera conditioning \textit{without} degrading synthesis quality.
\quad
To perform our analysis we train a \paramcount-parameter 
VDiT (video latent diffusion transformer)~\cite{DiT} on a dataset of 100M text/video pairs.
On this model, we perform \textit{three} key studies.
With what we learn, we adapt the camera control solution from VD3D~\citep{VD3D} from a pixel-based to latent-based diffusion model, and significantly improve its performance.

\paragraph{1) The spectral properties of camera motion}
To study the statistical nature of motion control, we analyze \emph{motion spectral volumes} (MSV)~\cite{GID} of the videos generated by a large-scale video DiT model.
MSVs show the amount of energy in different portions of the frequency spectra (i.e., high energy in the low frequencies indicate smooth motion) and we measure them across $200$ generated videos of different types (camera motion, scene motion, scene plus camera motion) and at various stages of the denoising synthesis process.
We observe that camera motion mostly affects the lower portion of the spectrum and kicks in very early (\apprx{10}\%) in the denoising trajectory.
Then, as diffusion models are inherently coarse-to-fine in nature~\cite{PerspectivesOnDiffusion}, we restrict our camera conditioning to only being injected on the subset of the denoising steps corresponding to low-frequencies.
This results in \apprx{15}\% higher visual fidelity, \apprx{30}\% better camera following, and mitigates scene motion degradation.

\paragraph{2) Camera motion knowledge in VDiTs}
Then, we consider our text-only \vdit, and determine \textit{whether} such a model possesses knowledge about cameras, and \textit{where} this knowledge is expressed within its  architecture.
With this objective, we feed the (\textit{unseen} during training) RealEstate10k~\cite{RealEstate10k} videos to our \vdit, and perform \textit{linear probing}~\citep{el2024probing} to determine if camera poses can be recovered from its internal representation.
Our analysis revealed that a video DiT implicitly performs camera pose estimation under the hood, and the presence of camera knowledge in a \emph{disentangled} form peaks in its middle layers.
This implies that the camera signal emerges in its early blocks to allow the later ones rely on it to build subsequent visual representations.
Therefore, we adjust our conditioning scheme to only affect the first $30\%$ of the architecture, leading to a \apprx{4}$\times$ reduction in training parameters, ${15}\%$ training and inference acceleration, and ${10}\%$ improved visual quality.

\paragraph{3) Re-balancing the training distribution}
Finally, to supervise camera control architectures, the typical solution is to rely on the camera pose annotations provided by RealEstate10k~\cite{RealEstate10k}.
However, this dataset contains mostly \emph{static} scenes, which results in significant motion degradation of the fine-tuned video model.
To overcome this problem, we curate a subset of $20$K diverse videos with dynamic scenes \emph{but} static cameras.
As the camera conditioning branch is still activated for these videos, this helps the model disambiguate the camera from scene movement.
Our experiments show that this simple adjustment in the data is sufficient to recover the scene dynamism while still enabling an effective pose-conditioned video model.

\paragraph{Contributions}
We compound the knowledge gained from these three studies into the design of the \methodfullname~(\methodname) method.
We perform extensive ablation studies and compare against state-of-the-art models for camera control, including MotionCtrl~\cite{MotionCtrl}, CameraCtrl~\cite{CameraCtrl}, and VD3D~\cite{VD3D}. 
We demonstrate $18\%$ higher video fidelity and $25\%$ more precise camera steering in terms of quantitative metrics than a closest competitor, and our generated videos are favored to others in \emph{$90\%$} of cases.

\section{Related work}
\label{sec:related-work}

Our approach lies at the intersection of text-to-video, text-to-3D, and text-to-4D generation approaches. We refer to recent state-of-the-reports~\cite{po2023state,yunus2024recent} for a more thorough analysis of previous work. 

\paragraph{Text-to-video generation}
Our approach builds on recent advancements in 2D video diffusion models. One prominent technique in this area enhances text-to-image models by adding temporal layers to support video generation~\citep{blattmann2023align,singer2022make,wu2023lamp,AnimateDiff,blattmann2023stable}. While these methods use the U-Net architecture, more recent ones~\citep{Sora,ma2024latte,SnapVideo, CogVideoX}
have been adapting transformer-based architectures for more scalable, realistic, and highly dynamic scene generation. 
We are interested in controlling the camera movements during the generation process of recent transformer-based video models based on precise camera extrinsics, i.e., cameras represented as rotation and translation sequences for each frame.

\begin{figure*}
\centering
\includegraphics[width=1.0\textwidth]{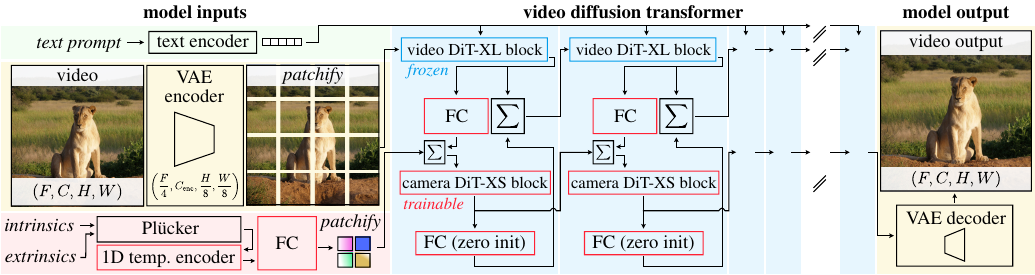}
\caption{\baselinewithref model with ControlNet~\cite{ControlNet,ControlNet-pp} camera conditioning built on top of~\vditwithref.
Video synthesis is performed by large 4,096-dimensional \texttt{DiT-XL} blocks of the frozen \vdit\ backbone, while \baselinename\ only processes and injects the camera information through lightweight 128-dimensional \texttt{DiT-XS} blocks (\texttt{FC} stands for fully-connected layers); see \Cref{sec:method:baseline-cc} for details.}
\label{fig:architecture}
\end{figure*}

\paragraph{4D generation}
Early 4D generation works~\citep{bahmani20223d,xu2022pv3d} used 4D GANs to learn category-specific generators with an underlying dynamic 3D representation. More recent approaches ~\citep{singer2023text,bahmani20234d,zheng2023unified, ling2023align,bahmani2024tc4d} have tackled 4D generation by distilling motion priors from pre-trained video diffusion models into an explicit 4D representation, enabling category-agnostic 4D generation. Follow-up works investigate image or video conditioned 4D generation ~\citep{ren2023dreamgaussian4d,zhao2023animate124,yin20234dgen,pan2024fast, zheng2023unified, ling2023align, gao2024gaussianflow, zeng2024stag4d, liang2024diffusion4d} instead of pure text inputs, improving flexibility in the generation process.
While most of these works are object-centric, recent approaches~\citep{bahmani2024tc4d, xu2024comp4d} shifted towards more complex scenes, including methods~\citep{chu2024dreamscene4d, yu20244real} which model the background. However, all these methods are optimization-based, i.e., each scene is generated independently from scratch. Recently, L4GM~\citep{ren2024l4gm} proposes a feed-forward 4D generator trained on object-centric synthetic 4D data.
While these approaches are explicit and provide space-time control, they are limited in their photorealism compared to recent 2D video diffusion models. We investigate dynamic 3D scene generation from a different perspective by extending pre-trained video diffusion models with 3D camera control.

\paragraph{Camera control for video models} 
Recently, there has been significant progress in adding camera control to video diffusion models. As the pioneering work, MotionCtrl~\citep{MotionCtrl} learns camera control by conditioning pre-trained video models~\citep{VideoCrafter1,blattmann2023stable} with extrinsics matrices. Follow-up works~\citep{CameraCtrl,xu2024camco, kuang2024collaborative} further improve the conditioning mechanisms by representing cameras as Plücker coordinates. Another line of work~\citep{MotionMaster,xiao2024video, ling2024motionclone,hou2024training} controls camera motion without training additional parameters. However, all of these approaches use U-Net-based architectures as their backbone.
More recently, 4DiM~\citep{4DiM} trains a space-time diffusion model from scratch for novel view synthesis from a single image input.
Closely related to our work, VD3D~\citep{VD3D} incorporates camera control into a pre-trained video diffusion transformer. While the motion and camera control improves over U-Net-based approaches, the synthesized motion in the scenes and the visual quality are still degraded compared to the base video model.
In contrast to VD3D, we first thoroughly investigate the pre-trained base video model and its knowledge of camera motion. We derive an improved training and architecture design for high-quality and dynamic video generation based on our findings.

\paragraph{Concurrent works} 
Concurrent approaches~\citep{zhao2024genxd, xu2024cavia, zheng2024cami2v, zhang2024recapture, yu2024viewcrafter,lei2024animateanything} further improve camera control in U-Net-based architectures, while another work~\citep{cheong2024boosting} tackles video diffusion transformer. However, the scene and visual quality is still limited in that approach.
DimensionX~\citep{sun2024dimensionx} controls space and time in video diffusion transformers but the camera trajectories are pre-defined and not continuous.
Chun-Hao et al.~\cite{CamPoseVDiT} explore pose estimation with a video DiT by pairing it with DUSt3R~\cite{dust3r} and fine-tuning, while we perform linear probing without any training to assess its existing camera knowledge. CAT4D~\cite{wu2024cat4d} proposes a multi-view video diffusion model fine-tuned from a multi-view model. By scaling multi-view synthetic data generation, SynCamMaster~\cite{bai2024syncammaster} and ReCamMaster~\cite{bai2025recammaster} achieve impressive results on camera-controlled video generation with transformers.

\section{Method}
\label{sec:method}
We first describe our base video diffusion model~(\cref{sec:method:base-video-dit}), and the baseline camera control method built on top of it~(\cref{sec:method:baseline-cc}).
Then, we proceed with the analysis of motion~(\cref{sec:method:motion}), linear probing~(\cref{sec:method:linprob}) and dataset biases~(\cref{sec:method:data}), and additional insights on how to build an effective model for camera control~(\cref{sec:method:misc})

\subsection{Base model~(\vdit)}
\label{sec:method:base-video-dit}
Following Sora~\cite{Sora}, most modern foundational text-to-video generators use the diffusion framework~\cite{DiffusionModels,DDPM} to train a large-scale transformer~\cite{Transformer} in the latent space of a variational autoencoder~\cite{VAE,LatentDiffusion}.
We adopt the same design and, for a base video model, pre-train an \paramcount-parameter Video DiT model~\cite{DiT} with 32 blocks of hidden dimension 4,096 for text-to-video generation. 
We use the rectified flow diffusion parametrization~\cite{RecFlow} and learn in the latent space of CogVideoX~\cite{CogVideoX} (using an autoencoder with a $16$-channel output and compression factors of $4{\times}8{\times}8$ in the temporal and spatial dimensions).
The T5~\cite{T5} encoder produces text embeddings, which are passed into \vdit\ via cross-attention.
We train our base model on a large-scale dataset of images and videos with text annotations, with resolutions ranging from $17{\times} 144 {\times} 256$ to $121 {\times} 576{\times}1024$.
This design is fairly standard and followed by many existing works with little deviation~\cite{MovieGen,CogVideoX,LuminaT2X,OpenSora}; we describe our specific architectural and training setup in detail in~\cref{supp:impl-details}.

\subsection{\vdit\ with Camera Control~(\baselinename)}
\label{sec:method:baseline-cc}
To construct a baseline architecture for camera control, we implement ControlNet~\cite{ControlNet,PixArt-delta} conditioning on top of the~\vdit.
Similar to previous work~\cite{MotionCtrl, CameraCtrl,VD3D}, we use the RealEstate10k~\cite{RealEstate10k} dataset, consisting of 65k (text, video, camera trajectory) triplets~${(\textcond_n, \video_n, \cameracond_n)}_{n=1}^N$ and train a new set of model parameters to input the camera information into the model.
Camera trajectories $\cameracond \in \R^{f\times25}$ are provided in the form of camera extrinsics $\camextr_f \in \R^{4\times 4}$ and intrinsics $\camintr_f \in \R^{3\times 3}$ for each $f$-th frame $\video_f$.

\paragraph{Camera conditioning}
\quad For base camera control, we adapt VD3D~\cite{VD3D} since it was designed for transformer-based models and suits our setup the most, while other methods are built on top of UNet-based~\cite{U-net} backbones.
We use \Plucker camera representations~\citep{sitzmann2021light,chen2023ray,kant2024spad,VD3D,CameraCtrl}, which are projected to the same dimensionality and resolution as the video tokens via a fully-convolutional encoder to produce camera tokens.
\quad These camera tokens are processed by a sequence of lightweight \ditxs\ blocks with hidden dimension 128 and four attention heads each.
To mix the camera information with the video tokens of \vdit, we use summation before each main DiT block.
We also found it useful to perform cross-attention from video tokens to camera tokens as a form of a feedback connection~\cite{ControlNet-pp}.
We illustrate this model architecture, which we call \baselinename, in \Cref{fig:architecture}; see implementation details in \Cref{supp:impl-details}. \baselinename describes the camera-controlled video model architecture used by AC3D, while AC3D describes our proposed work including analysis and additional adjustments based on the analysis.

\paragraph{Training}
Keeping the \vdit backbone frozen, we train the new parameters with a rectified flow objective~\cite{RecFlow} and standard (location of 0 and scale of 1) logit-normal noise distribution~\cite{SD3}.
Similar to prior works~\cite{VD3D, 4DiM}, we apply a $10\%$ camera dropout to support classifier-free guidance~(CFG)~\cite{CFG} later.
Notably, we train \baselinename only at the $256^2$ resolution: since camera motion is a low-frequency type of signal (which can be observed at lower resolutions) and the main \vdit backbone is frozen, we found that our design generalizes to higher resolutions out-of-the-box.
During inference, we input text prompts and camera embeddings with classifier-free guidance at each time step.

\begin{figure}
\centering
\includegraphics[width=0.8\linewidth]{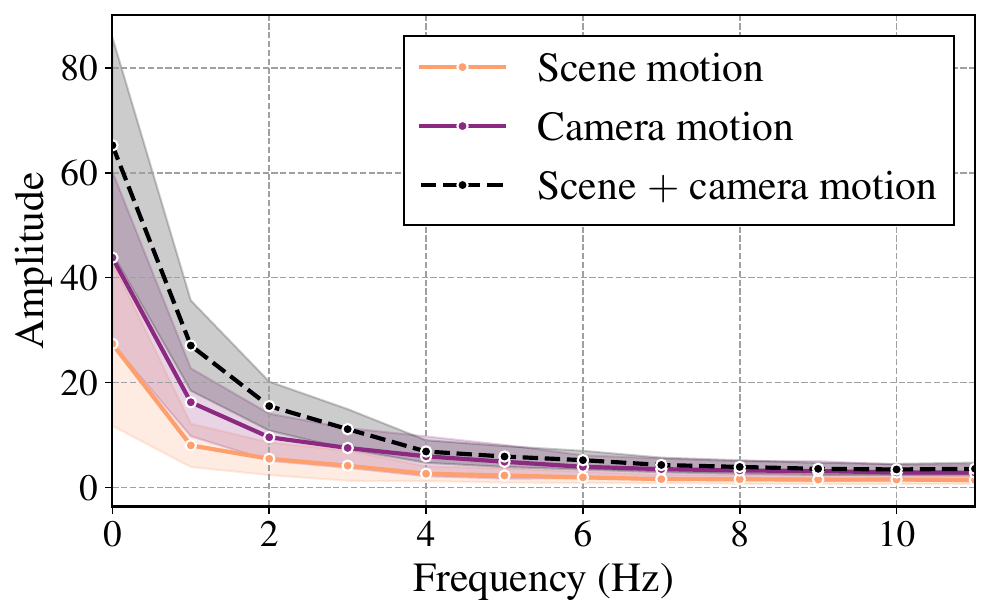}
\caption{
Average magnitude of motion spectral volumes along spatial, temporal offset, and video batch dimensions for scenes with different motion types. We compute the flow of each video in a sliding window manner with temporal offsets and average the frequencies across all offsets.
Videos with camera motion (purple) exhibit stronger overall motion than the videos with scene motion (orange), especially for the low-frequency range, suggesting that the motion induced by camera transitions is heavily biased towards low-frequency components. Frequency refers to the temporal frequency.
}
\label{fig:spec-comparison}
\end{figure}

\paragraph{Model behavior}
This baseline model, being built on top of our powerful \vdit, already achieves decent-quality camera control.
However, it struggles with degraded visual quality and reduced scene motion, and sometimes, the camera control inputs are ignored.
To improve the design, we analyze our \vdit\ backbone to understand how camera motion is modeled and represented. 
Then, we inspect \baselinename's failure cases and where they arise to address them.

\subsection{How is camera motion modeled by diffusion?}
\label{sec:method:motion}
We start by analyzing how camera motion is modeled by a pre-trained video diffusion model (i.e., before camera control is incorporated).
We hypothesize that the motion induced by changes in camera pose is a low-frequency type of signal and investigate the \textit{motion spectral volumes}~\cite{GID} of the generated videos at different steps of the denoising process.
To perform this analysis, we generate $200$ diverse videos with our \vditwithref{} model with 80 denoising steps and \textit{manually} annotate them into four categories: videos with only scene motion, videos with only camera motion, videos with both scene and camera motion, and others; see \Cref{supp:motion-analysis-details} for details.
During generation, we save the denoised predictions at \textit{each} denoising step and estimate optical flow to compute the motion spectral volumes.

\begin{figure*}[t]
\centering
\begin{subfigure}[t]{0.32\linewidth}
\vspace{0.05cm}
\includegraphics[width=\linewidth]{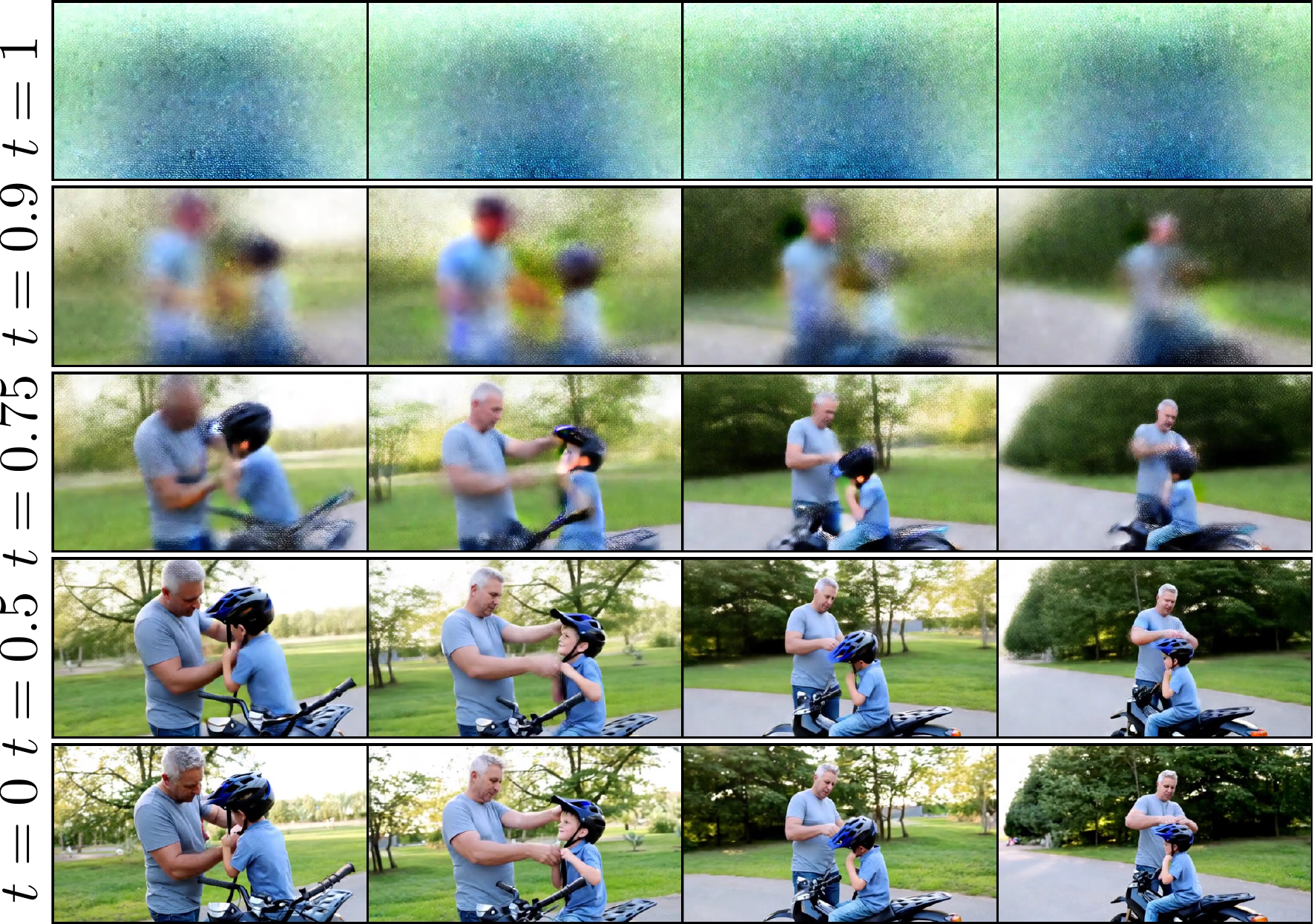}
\caption{A generated video at different diffusion timesteps. The camera has already been decided by the model even at $t = 0.9$ (first 10\% of the denoising process) and does not change after that.}
\label{fig:spec-analysis:visual-example}
\end{subfigure}%
\hfill
\begin{subfigure}[t]{0.66\linewidth}
\vspace{0pt}
\includegraphics[width=\linewidth]{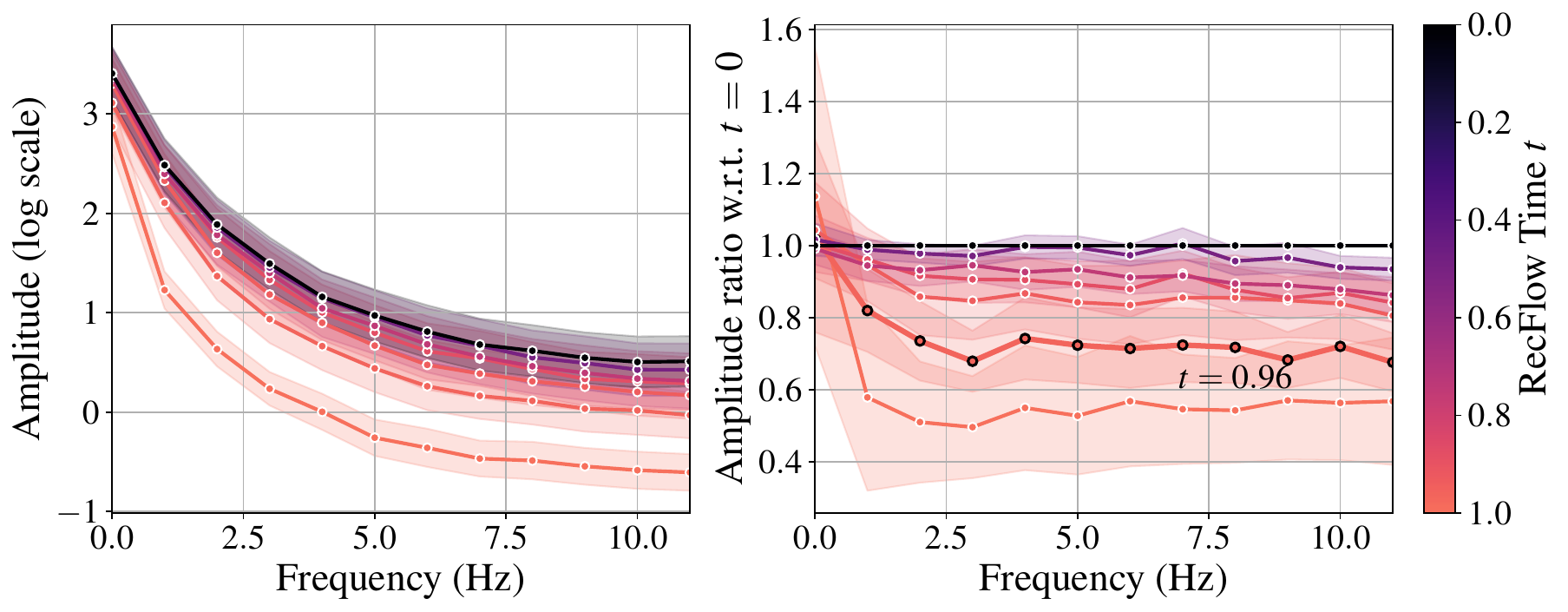}
\caption{Motion spectral volumes of \vditwithref's generated videos for different diffusion timesteps (left) and their ratio w.r.t. the motion spectral volume at $t=0$ (i.e., a fully denoised video).}
\label{fig:spec-analysis:spec-per-timestep}
\end{subfigure}%
\caption{
\textbf{How camera motion is modeled by diffusion?}
As visualized in \Cref{fig:spec-analysis:visual-example} and \Cref{fig:spec-comparison}, the motion induced by camera transitions is a low-frequency type of motion.
We observe that a video DiT creates low-frequency motion very early in the denoising trajectory: \Cref{fig:spec-analysis:spec-per-timestep} (left) shows that even at $t{=}0.96$ (first \apprx{4}\% of the steps), the low-frequency motion components have already been created, while high frequency ones do not fully unveil even till $t{=}0.5$.
We found that controlling the camera pose later in the denoising trajectory is not only unnecessary but \emph{detrimental} to both scene motion and overall visual quality.
}
\label{fig:spec-analysis}
\end{figure*}

\paragraph{Analysis}
We visualize motion spectral volumes with 95\% confidence intervals in \Cref{fig:spec-comparison}.
Videos with camera motion exhibit higher amplitudes than scene-motion-only videos for low-frequency components while having similar characteristics for high-frequency ones.
This supports the conjecture that the camera motion is a low-frequency type of signal.
We also depict an example of a generated video with both scene and camera motion with four denoising steps on \cref{fig:spec-analysis:visual-example}: one can observe that the camera movement has been fully produced by $t{=}0.9$ (first $10\%$ of the rectified flow denoising process). In contrast, scene motion details like the hand movements of the subjects are not finalized even till 
$t{=}0.5$.
\par Inspired by this finding, we pose the question: \textit{when exactly does a video diffusion model determine the camera pose?}
To answer this question, we plot aggregated spectral volumes for different timesteps in \Cref{fig:spec-analysis:spec-per-timestep}. 
We also show the ratio with respect to the last timestep $t{=}0$ (i.e., when all motion has been generated).
We then inspect when different types of motion appear during the denoising process.
\Cref{fig:spec-analysis:spec-per-timestep}~(right) shows that the low-frequency motion components fill up to \apprx{84}\% at $t{=}0.9$ (the first 10\% of the denoising process), while high-frequency components are not well-modeled until $t{=}0.6$.
\par An immediate consequence of this observation is that trying to control the camera later in the denoising trajectory is simply \textit{unnecessary} and will not influence the manipulation result.
In this way, instead of using the standard logit-normal noise level distribution of SD3~\cite{SD3} with a location of 0.0 and scale of 1.0 (which we use by default for \vdit), we switch to using truncated normal with a location of 0.8 and scale of 0.075 on the [0.6, 1] interval to cover the early steps of the denoising rectified flow trajectory.
At inference time, we apply camera conditioning on the same [0.6, 1] interval.
Surprisingly, we observe that not using truncation is \textit{detrimental} to the scene motion and overall visual quality.
Following this insight, we restrict both our train-time noise levels and test-time camera conditioning schedules to cover only the first 40\% of the reverse diffusion trajectory.
As \cref{sec:experiments:ablations} shows, this improves FID and FVD by $14\%$ on average, and camera following by $30\%$ on MSR-VTT (the dataset used to measure generalization to diverse, out-of-fine-tuning-distribution scenes).
Further, truncated noise sampling enhances the overall scene motion. 

\begin{figure*}
\centering
\includegraphics[width=0.95\linewidth]{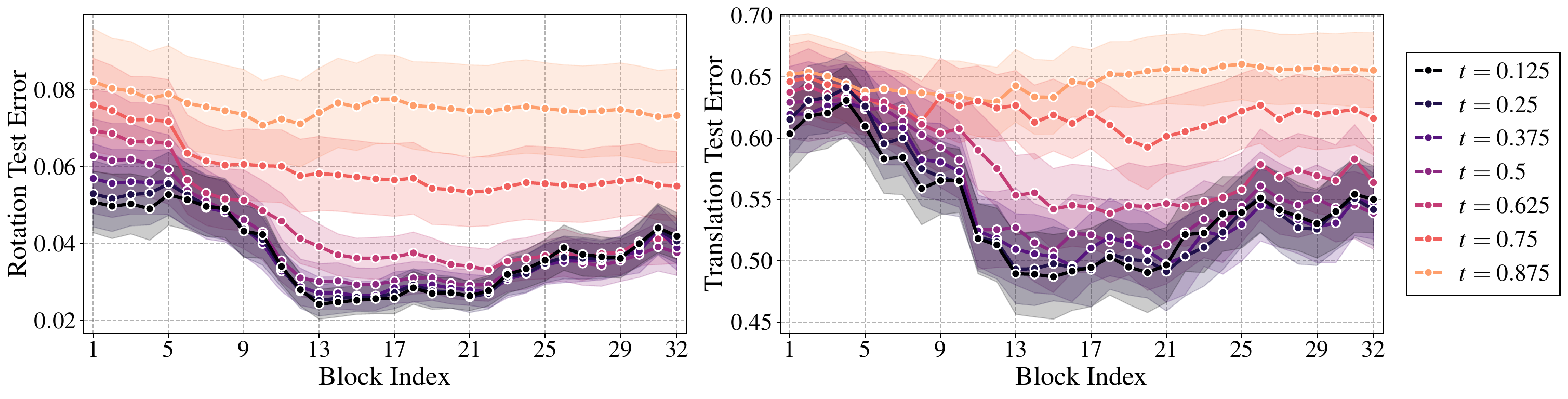}
\caption{
\textbf{Video DiT is secretly a camera pose estimator.}
We perform linear probing of camera poses in each of \vditwithref\ blocks for various noise levels and observe that video DiT performs pose estimation under the hood.
Its middle blocks carry the most accurate information about the camera locations and orientations, which indicates that the camera signal emerges in the early layers to help the middle and late blocks render other visual features aligned with the viewpoint.
}
\label{fig:linprob-analysis}
\end{figure*}

\subsection{What does \vdit know about camera pose?}
\label{sec:method:linprob}
\label{sec:method:knowledge}
Foundational video models acquire rich knowledge about the physical world, and we hypothesize that they store information about the camera pose within their representations.
To investigate this, we perform linear probing of our base \vditwithref\ model on the RealEstate10k~\cite{RealEstate10k} dataset (not seen during training) for camera extrinsics.
\par Specifically, we take 1,000 random 49-frame videos from RealEstate10K, feed them into \vdit\ under 8 noise levels $(1/8, 2/8, ..., 1)$, and extract the activations for all 32 DiT blocks.
Next, we split the random videos into 900 train and 100 test videos and train a linear ridge regression model to predict the rotation pitch/yaw/roll angles and translation vectors for the entire viewpoint trajectory ($49\times 6$ target values in total).
This results in $8 \times 32$ trained models, and we report the rotation and (normalized) translation errors~\cite{CameraCtrl} on a held-out test set of 100 videos in \Cref{fig:linprob-analysis}.
Surprisingly, \vdit can accurately predict the camera pose, achieving minimum test errors of \apprx{0.025} for rotation and for \apprx{0.48} translation prediction.
The knowledge quality increases around layer \#9 and peaks in the range of \#13--21.
We reason that since the camera information in block \#13 is stored in such a disentangled manner, then the model is using it to build other representations; hence, conditioning the camera in this block is risky and unnecessary and would interfere with other visual features, as shown in our ablations.
In this way, we propose to input the camera conditioning \textit{only} in the first \#8 blocks and leave the remaining 24 DiT blocks unconditioned.
We find in \Cref{sec:experiments:ablations} that this not only reduces the number of trainable parameters by \apprx{4} times and improves training speed by \apprx{\linprobtrainspeedup}\%, but also enhances the visual quality by \apprx{10}\%.

\subsection{Mitigating training data limitations}
\label{sec:method:data}
Estimating camera parameters from in-the-wild videos remains challenging, as leading methods like~\cite{particlesfm, colmap_mvs, colmap_sfm, dust3r} frequently fail when processing videos containing dynamic scene content.
This limitation results in camera-annotated datasets being heavily biased toward static scenes, which is particularly evident in \retenkfull~(\retenk)~\cite{RealEstate10k}, the predominant dataset for training camera-controlled video models~\cite{MotionCtrl,CameraCtrl,VD3D}.
We hypothesize that models fine-tuned on such data interpret camera position information as a signal to suppress scene dynamics.
This bias persists even when jointly training on unconstrained 2D video data~\cite{4DiM}, because the camera conditioning branch is only activated when camera parameters are available, which occurs exclusively for static scenes from \retenk, as static scenes remain the only reliable source for accurate camera annotation.
\par To address this fundamental limitation, we propose an alternative approach: rather than attempting to annotate dynamic scenes, which proved unsuccessful in our extensive preliminary research, even with state-of-the-art methods~\cite{dust3r}, we curate a collection of $20$K diverse videos featuring dynamic scenes captured by stationary cameras (see~\Cref{fig:getty-static}).
With stationary cameras, the camera position is inherently known, (we can assign fixed arbitrary extrinsics), allowing us to maintain active camera conditioning during training.
This approach enables the camera conditioning branch to remain active during training while exposing the model to dynamic content, helping it distinguish between viewpoint conditioning and scene stillness.
On top of this secondary dataset, following~\cite{4DiM}, we remove the scale ambiguity in \retenk by leveraging an off-the-shelf metric depth estimator; see~\Cref{supp:scaling-bias}.
Our experiments in \cref{sec:experiments:ablations} demonstrate that this straightforward yet effective data curation strategy successfully mitigates the distributional limitations of \retenk, restoring much of the lost scene dynamics, while maintaining precise camera control.

\newcommand{\gtstaticwidth}{0.93\linewidth}
\begin{figure}
\centering
\begin{tabular}{@{}c@{\hspace{0.3em}}l@{}}
\rotatebox[origin=c]{90}{\small{\retenkfull}} &
\begin{tabular}{@{}c@{}}
    \includegraphics[width=\gtstaticwidth]{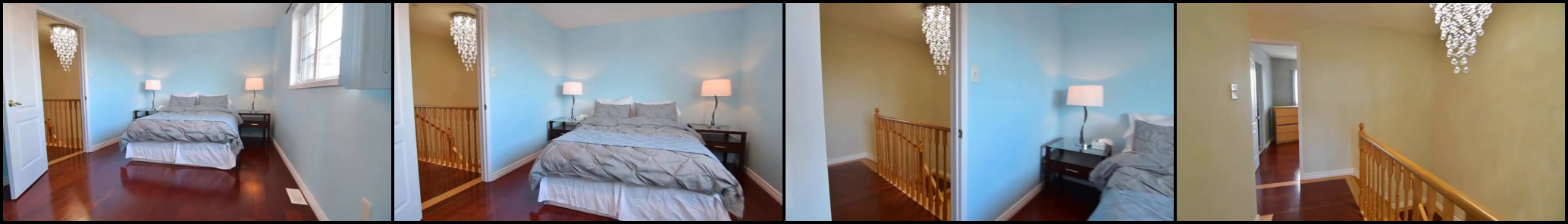}\\
    \includegraphics[width=\gtstaticwidth]{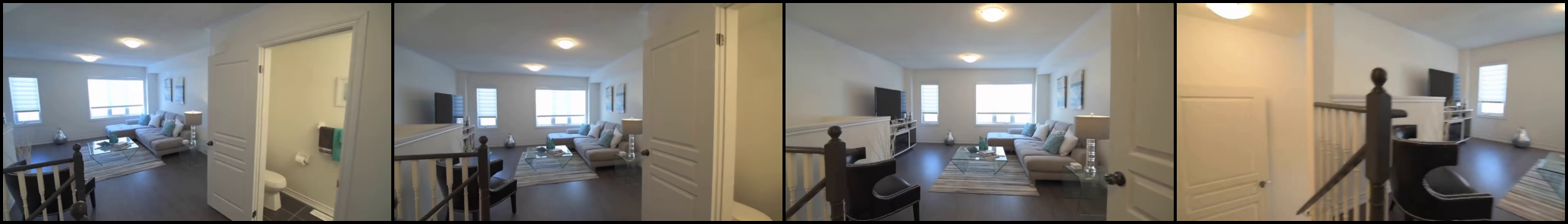}
\end{tabular}\\[0.3em]
\rotatebox[origin=c]{90}{\small{Our curated data}} &
\begin{tabular}{@{}c@{}}
    \includegraphics[width=\gtstaticwidth]{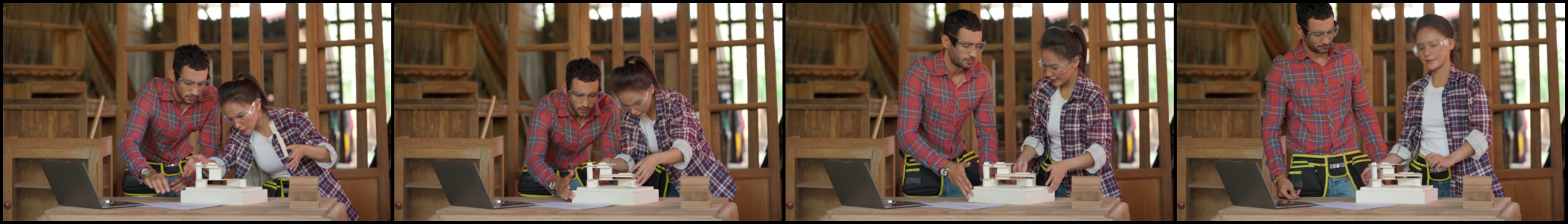}\\
    \includegraphics[width=\gtstaticwidth]{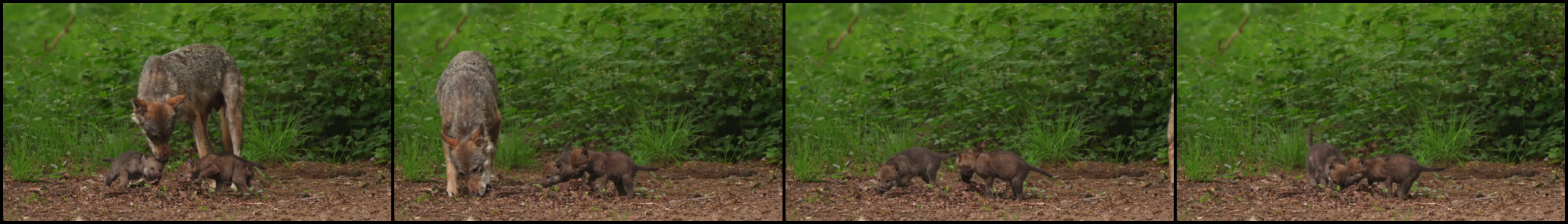}
\end{tabular}
\end{tabular}
\caption{
RealEstate10k~\cite{RealEstate10k} videos (upper 2 rows) contain diverse camera trajectories, but are strongly biased towards static scenes.
To mitigate this bias and also increase the concepts diversity, we curate $20$K videos with stationary cameras, but dynamic content (lower 2 rows).
Such datasets are easy to construct, and surprisingly effective. \Cref{sec:experiments:ablations} shows that integrating the dataset into our training improves visual quality on out-of-distribution prompts by $17\%$.
}
\label{fig:getty-static}
\end{figure}

\subsection{Miscellaneous improvements}
\label{sec:method:misc}
In addition to our core analysis, we introduce several auxiliary techniques that enhance model performance.

\paragraph{Separate text and camera guidance}
Text and camera signals require different guidance weights due to their distinct nature, motivating us to separate their classifier-free guidance (CFG)~\citep{CFG,InstructPix2Pix}.
We formulate the equation as:
\begin{align}
\hat{s}(\bm x | \textcond, \cameracond) 
&= (1 + w_y + w_c) s_\theta(\video | \textcond, \cameracond)  \\
&- w_y s_\theta(\video | \cameracond) - w_c s_\theta(\video | \textcond), \nonumber
\label{eq:separate-cfg}
\end{align}
where $\hat{s}(.)$ denotes the final update direction used during synthesis, $s_\theta$ represents the model's predicted update direction, $\textcond$ and $\cameracond$ are text and camera conditions, and $w_y$ and $w_c$ are their respective CFG weights. We zero-out the tensor for unconditional generation.

\paragraph{ControlNet with feedback}
Traditional ControlNet~\cite{ControlNet} conditioning, used in recent camera control methods~\cite{MotionCtrl,CameraCtrl,VD3D}, only processes conditioning signals without accessing the main branch.
Our experiments show that using a bidirectional ControlNet produces better camera representations. 
This modification behaves as a feedback mechanism~\cite{ControlNet-pp} provided by the main synthesis branch to the camera processing branch.

\paragraph{Dropping context in the camera branch}
Applying cross-attention over the context information (text prompts, resolution, etc.) in the \camditxs\ blocks worsens visual quality and camera steering due to harmful interference of the context embeddings with camera representations.

\section{Experiments}
\label{sec:experiments}

\paragraph{Datasets}
Our base \vditwithref\ model was trained on a large-scale dataset of text-annotated images and videos.
\baselinewithref\ is fine-tuned from \vdit\ on \retenkfull~\cite{RealEstate10k}, contains \apprx{65}M video clips with per-frame camera parameters since it is the setup used by existing methods~\cite{MotionCtrl, CameraCtrl, VD3D}.

\paragraph{Metrics}
To assess the performance, we rely on a wide range of automatic quantitative metrics.
We use FID~\cite{FID}, FVD~\cite{FVD}, and CLIP score~\cite{CLIPScore} to evaluate visual quality, and rotation and normalized translation errors~\cite{CameraCtrl} of ParticleSfM~\cite{particlesfm}-reconstructed trajectories to assess camera steerability.
We evaluate them both on \retenk\ and \msrvtt~\cite{MSRVTT}, since the latter allows to assess zero-shot generalization on out-of-distribution data. Moreover, we conduct a user study with details in the appendix in Sec.~\ref{supp:user_study}.

\begin{table}[t]
\caption{\textbf{User study.} We compare our approach to the original VD3D (FIT) and reimplemented VD3D (DiT) on top of our base model. We conduct a user study where participants indicate their preference based on \textit{camera aligntment (CA)}, \textit{motion quality (MQ)}, \textit{text alignment (TA)}, \textit{visual quality (VQ)}, \textit{and overall preference~(Overall)}.}
\label{tab:user_study}
\begin{center}
\resizebox{1.0\textwidth}{!}{
\begin{tabular}{l|cccc|c}
    \toprule
     &  \multicolumn{5}{c}{Human Preference}\\
    Method & \textit{CA} & \textit{MQ} & \textit{TA} & \textit{VQ} & \textit{Overall} \\\midrule
    Ours vs. VD3D (FIT) & 89.5\% & 79.0\% & 87.5\% & 97.5\% & 95.0\%\\
    Ours vs. VD3D (DiT) & 65.0\% & 87.5\% & 83.5\% & 95.0\% & 92.5\%\\
    \bottomrule
\end{tabular}}
\end{center}
\end{table}
\subsection{Baselines}
\label{sec:baselines}

We select three camera-control methods: MotionCtrl~\cite{MotionCtrl}, CameraCtrl~\cite{CameraCtrl}, and VD3D~\cite{VD3D}.
MotionCtrl and CameraCtrl use a UNet-based video diffusion backbone~\cite{AnimateDiff}, while VD3D builds on top of FIT~\cite{SnapVideo,FIT} and as such, is easily extendable to our video DiT~\cite{DiT} setup.
Hence, we re-implement VD3D on top of our \vditwithref\ model to obtain an additional ``VD3D+DiT'' baseline.
Moreover, we provide comparisons for an open-source model, i.e., CogVideoX~\cite{yang2024cogvideox}. See Sec.~\ref{supp:cogvideox} of the appendix for more details.

\newcommand{\transerrorname}{TransErr $\downarrow$}
\newcommand{\roterrorname}{RotErr $\downarrow$}
\newcommand{\fidname}{FID $\downarrow$}
\newcommand{\fvdname}{FVD $\downarrow$}
\newcommand{\clipscorename}{CLIP $\uparrow$}

\begin{table*}[!t]
\setlength{\tabcolsep}{2pt}
\renewcommand{\arraystretch}{0.9}
\caption{\textbf{Quantitative evaluation.} We evaluate all the models using camera pose and visual quality metrics based on unseen camera trajectories. We compute translation and rotation errors based on the estimated camera poses from generations using ParticleSfM \citep{particlesfm}. We evaluate both in-distribution performance with RealEstate10K~\citep{RealEstate10k} and out-of-distribtion performance with MSR-VTT~\cite{MSRVTT}. 
}
\label{table:metrics_camera_pose}

\footnotesize
\begin{tabular*}{1.0\textwidth}{@{\hspace{1pt}}l@{\extracolsep{\fill}}cccccccccc}
\toprule
\multirow{2}{*}{Method} & \multicolumn{5}{c}{RealEstate10K~\cite{RealEstate10k}} & \multicolumn{5}{c}{MSR-VTT~\cite{MSRVTT}} \\
\cmidrule(lr){2-6} \cmidrule(lr){7-11}
& \transerrorname & \roterrorname & \fidname & \fvdname & \clipscorename & \transerrorname & \roterrorname & \fidname & \fvdname & \clipscorename \\
\midrule
MotionCtrl (U-Net) & 0.477 & 0.094 & 2.99 & 61.70 & 26.46 & 0.593 & 0.137 & 16.85 & 283.12 & 24.11 \\
CameraCtrl (U-Net) & 0.465 & 0.089 & 2.48 & 55.64 & 26.81 & 0.587 & 0.132 & 12.33 & 201.33 & 25.05 \\
VD3D (FIT) & 0.409 & \textnormal{0.043} & 1.40 & 42.43 & 28.07 & 0.504 & 0.050 & 7.80 & 165.18 & 26.89 \\
\midrule
VD3D (CogVideoX) & 0.467 & 0.063 & 1.66 & 43.14 & 28.08 & 0.501 & 0.068 & 7.45 & 148.11 & 27.65 \\
\methodname (CogVideoX)~\ours & \cellsecond{0.374} & \cellsecond{0.039} & \textnormal{1.27} & \cellsecond{38.20} & \cellsecond{28.62} & \cellsecond{0.431} & \cellsecond{0.039} & \cellsecond{5.52} & \cellsecond{116.04} & \cellsecond{28.38} \\
\midrule
MotionCtrl (VDiT) & 0.504 & 0.126 & 1.74 & 43.81 & 27.69 & 0.589 & 0.146 & 9.92 & 150.20 & 27.25 \\
CameraCtrl (VDiT) & 0.513 & 0.138 & 1.62 & 42.10 & 27.73 & 0.566 & 0.143 & 8.15 & 146.77 & 27.51 \\
VD3D (VDiT) & 0.421 & 0.056 & \cellsecond{1.21} & \textnormal{38.57} & \textnormal{28.34} & \textnormal{0.486} & \textnormal{0.047} & \textnormal{6.88} & \textnormal{137.62} & \textnormal{27.90} \\
\methodname (VDiT)~\ours & \cellbest{0.358} & \cellbest{0.035} & \cellbest{1.18} & \cellbest{36.55} & \cellbest{28.76} & \cellbest{0.428} & \cellbest{0.038} & \cellbest{5.34} & \cellbest{110.71} & \cellbest{28.58} \\
\midrule
w/o camera cond &\cellcolor{Red!30}+0.233 &\cellcolor{Red!30}+0.153 &\cellcolor{Red!30}+4.02 &\cellcolor{Red!30}+53.83 &\cellcolor{Red!30}-1.63 &\cellcolor{Red!30}+0.266 &\cellcolor{Red!30}+0.157 &\cellcolor{Green!30}-0.48 &\cellcolor{Green!30}-8.53 &\cellcolor{Green!30}+0.35 \\
w/o biasing noise &\cellcolor{Red!30}+0.093 &\cellcolor{Red!9}+0.015 &\cellcolor{Red!0}+0.02 &\cellcolor{Red!3}+1.78 &\cellcolor{Red!19}-0.32 &\cellcolor{Red!30}+0.138 &\cellcolor{Red!21}+0.033 &\cellcolor{Red!30}+0.59 &\cellcolor{Red!30}+16.92 &\cellcolor{Red!30}-0.54 \\
w/o noise truncation &\cellcolor{Red!8}+0.020 &\cellcolor{Green!1}-0.003 &\cellcolor{Red!1}+0.06 &\cellcolor{Red!3}+1.69 &\cellcolor{Red!12}-0.20 &\cellcolor{Red!6}+0.016 &\cellcolor{Red!3}+0.005 &\cellcolor{Red!30}+0.76 &\cellcolor{Red!30}+6.63 &\cellcolor{Red!30}-0.18 \\
w/o camera CFG &\cellcolor{Red!6}+0.014 &\cellcolor{Red!2}+0.004 &\cellcolor{Red!12}+0.49 &\cellcolor{Red!8}+4.57 &\cellcolor{Red!30}-0.54 &\cellcolor{Red!9}+0.025 &\cellcolor{Red!1}+0.003 &\cellcolor{Red!3}+0.03 &\cellcolor{Red!8}+1.42 &\cellcolor{Red!30}-0.27 \\
w/o our dynamic data &\cellcolor{Green!2}-0.005 &\cellcolor{Green!2}-0.004 &\cellcolor{Green!1}-0.06 &\cellcolor{Red!0}+0.22 &\cellcolor{Red!12}-0.20 &\cellcolor{Red!1}+0.004 &\cellcolor{Green!0}-0.001 &\cellcolor{Red!30}+0.89 &\cellcolor{Red!26}+4.40 &\cellcolor{Red!30}-0.55 \\
w/o metric scaled data &\cellcolor{Red!5}+0.013 &\cellcolor{Red!3}+0.005 &\cellcolor{Red!4}+0.17 &\cellcolor{Red!8}+4.65 &\cellcolor{Red!0}0.00 &\cellcolor{Red!8}+0.023 &\cellcolor{Red!1}+0.002 &\cellcolor{Green!1}-0.01 &\cellcolor{Red!0}0.00 &\cellcolor{Red!21}-0.12 \\
w/o dropping camera context &\cellcolor{Red!5}+0.013 &\cellcolor{Red!0}+0.001 &\cellcolor{Red!0}+0.04 &\cellcolor{Red!4}+2.46 &\cellcolor{Red!30}-0.65 &\cellcolor{Red!10}+0.029 &\cellcolor{Red!1}+0.003 &\cellcolor{Red!30}+1.25 &\cellcolor{Red!30}+7.41 &\cellcolor{Red!30}-0.36 \\
w/o limiting camera cond to 8 blocks &\cellcolor{Green!0}-0.001 &\cellcolor{Red!0}+0.001 &\cellcolor{Red!2}+0.09 &\cellcolor{Red!1}+0.56 &\cellcolor{Red!1}-0.02 &\cellcolor{Red!1}+0.003 &\cellcolor{Red!0}0.000 &\cellcolor{Red!30}+0.32 &\cellcolor{Red!30}+9.23 &\cellcolor{Red!30}-0.33 \\
w/ 2D training &\cellcolor{Red!30}+0.129 &\cellcolor{Red!30}+0.068 &\cellcolor{Red!30}+2.60 &\cellcolor{Red!30}+33.85 &\cellcolor{Red!30}-1.17 &\cellcolor{Red!30}+0.128 &\cellcolor{Red!30}+0.093 &\cellcolor{Green!29}-0.26 &\cellcolor{Green!22}-3.83 &\cellcolor{Green!30}+0.21 \\
\bottomrule
\end{tabular*}
\end{table*}

\subsection{Main results}
\label{sec:comparisons}

We present quantitative comparisons with the baselines in \cref{table:metrics_camera_pose}.
One can observe that just switching from the 4B-parameter pixel-space FIT~\cite{FIT} backbone, employed by the original VD3D approach, to our larger \paramcount-parameter latent-space DiT yields clear improvements across most metrics.
Next, the results demonstrate that \methodname\ establishes a new state-of-the-art in performance against all baselines. 
Evaluating the quality of camera motion from still images is difficult, so we instead visualize all qualitative results in the website provided within our supplementary material.
Therein, we can observe that \methodname\ better follows pose conditioning and achieves higher visual fidelity.
We conduct user studies against VD3D+FIT (the original model) and VD3D+DiT (our improved re-implementation on top of the bigger video transformer).
The results are presented in \Cref{tab:user_study}: \methodname\ outperforms them across all qualitative aspects, achieving a 90\%+ overall preference score.
Finally, we encourage the reader to assess the visual quality by observing videos on our website.

\subsection{Ablations}
\label{sec:experiments:ablations}

\paragraph{No camera conditioning}
The first ablation we conduct is to drop all camera conditioning, which makes the model equivalent to the vanilla \vditwithref.
This is needed to understand the degradation of visual quality and text alignment.
The results (\cref{table:metrics_camera_pose}, row \abl{w/o camera cond}) show that our model loses less only \apprx{7}\% of the original visual fidelity on \msrvtt~(as measured by FVD), while (as expected) greatly improving on its in-domain \retenk data.
In comparison, VD3D-DiT (the closest baseline) loses \apprx{20}\% of its visual fidelity on \msrvtt.

\paragraph{Importance of biasing the noise towards higher levels}
As \cref{sec:method:motion} shows, we use the truncated normal distribution with location of 0.8 and scale of 0.075 with the [0.6, 1] bounds for training \methodname.
We ablate the importance of biasing the noise sampling towards high noise and observe higher motion, visual quality, and camera controllability. 

\paragraph{Importance of truncating the noise schedule}
We change the training and inference procedure by using no truncation during noise sampling. Instead, we condition the model with camera inputs over the whole noise range and observe decreased visual quality.

\paragraph{No camera guidance}
We assess the importance of classifier-free guidance~\cite{CFG} on the camera conditioning in \cref{table:metrics_camera_pose} (\abl{w/o camera CFG}).
It attains the same visual quality on both in-distribution (\retenk) and out-of-distribution (\msrvtt) data, but degrades camera following, resulting in \apprx{5}\% worse pose reconstruction errors.

\paragraph{Training without our data with scene motion}
To understand how well our curated data with scene motion but stationary cameras mitigates static scene bias, we train \methodname\ exclusively on \retenk, and report the results in \cref{table:metrics_camera_pose} (\abl{w/o our dynamic data}).
The model maintains similar visual quality and text alignment on \retenk (in-domain data), but performance on out-of-distribution samples from \msrvtt\ worsens (\apprx{17}\% worse FID and \apprx{4}\% worse FVD).
The quality of scene motion is better assessed by referring to our qualitative video comparisons in the supplementary. 

\paragraph{Importance of metric scaled cameras}
We train \methodname\ using the original \retenk's camera parameters without our scaling procedure and present the results in \cref{table:metrics_camera_pose} (\abl{w/o metric scaled data}).
This is a more ambiguous conditioning signal, and results in worse visual quality (\apprx{10}\% FVD on \retenk) and camera following performance (\apprx{12}\% worse trajectory reconstruction).

\paragraph{Providing context into the camera branch}
As discussed in \cref{sec:method:misc}, we chose not to input the context information (text embeddings, resolution conditioning, etc.) into the camera branch to avoid potential interference with the camera representations.
As \cref{table:metrics_camera_pose} (\abl{w/o dropping camera context}) shows, providing this information indeed results in \apprx{4}\% worse camera following and \apprx{15}\% lower visual quality.

\paragraph{Importance of limiting conditioning to the first 8 \vdit\ blocks}
Following our insights in \cref{sec:method:linprob}, we condition \methodname\ only in the first 8 blocks.
Trying to condition in all the 32 DiT blocks (\abl{w/o limiting camera cond to 8 blocks}) worsens the visual quality by \apprx{10}\%, while keeping the quality control at the same level.
This suggests that the middle and late \vdit\ layers indeed rely on processed camera information and conditioning them on external camera poses might lead to interference with other visual features.

\paragraph{Joint training with 2D data}
To mitigate visual quality and scene motion degradation, we attempted to perform joint fine-tuning on 2D video data (without camera annotations) which was used in base \vdit\ training by applying dropout on camera inputs for it.
Prior work shows performance benefits with this strategy~\cite{4DiM} and, as \cref{table:metrics_camera_pose} (\abl{with 2D training}) shows, it indeed helps to maintain slightly higher visual fidelity in our case (\apprx{3}\% better FVD on \msrvtt). 
However, camera steering severely deteriorates, leading to up to $3\times$ worse results for translation/rotation errors.

\section{Conclusions}
\label{sec:conclusuion}
Our findings demonstrate that principled analysis of camera motion in video diffusion models leads to significant improvements in control precision and efficiency.
Through enhanced conditioning schedules, targeted layer-specific camera control, and better-calibrated training data, \methodname\ achieves state-of-the-art performance in 3D camera-controlled video synthesis while maintaining high visual quality and natural scene dynamics.
This work establishes a foundation for more precise and efficient camera control in text-to-video generation.
We discuss the limitations of our approach 
in \cref{supp:limitations}.
In future work, we plan to focus on further improving data limitations and developing control mechanisms for camera trajectories far outside of the training distribution.

\section{Acknowledgements}
DBL acknowledges support from NSERC under the RGPIN program, the Canada Foundation for Innovation, and the Ontario Research Fund.

{
    \small
    \bibliographystyle{ieeenat_fullname}
    \bibliography{main}
}


\clearpage
\maketitlesupplementary
\appendix

\emph{We encourage the reader to inspect our visual results, comparisons with other models and additional visualizations in the accompanying website in \url{https://snap-research.github.io/ac3d}}.

\section{Ethics Statement}
\label{supp:ethics-statement}
As with all generative AI technologies, there is the potential for misuse by bad actors.
However, we anticipate this technology will advance creative expression, education, and research through:
\begin{enumerate}
    \item Enhanced creative tools enabling filmmakers and educators to achieve complex camera movements without specialized equipment, democratizing high-quality video production and expanding possibilities for visual storytelling.
    \item More realistic synthetic video that better simulates real world camera behaviors, improving applications in training autonomous systems, virtual production, and educational simulations where accurate camera dynamics are crucial.
    \item Advancing our technical understanding of how camera motion affects visual perception and generation, contributing to fundamental research in computer vision, graphics, and
human visual processing.
\end{enumerate}
\section{Limitations}
\label{supp:limitations}

In our work, we substantially advance the quality of 3D camera control of video diffusion models, but our analysis and method are not free from limitations.

\paragraph{OOD trajectories generalization}
Both our model and all the baselines struggle to generalize to the camera trajectories, which are far away from the training distribution of RealEstate10K~\cite{RealEstate10k}.
While, in general, it is an expected behavior, it indicates that the model processes the viewpoint conditioning information in a way that is entangled with the main video representations.
Another source of this issue is the pre-training distribution of the base \vdit itself: natural videos typically have simple recording trajectories and rarely exhibit something that looks like 3D scanning.
In this way, producing OOD trajectories is not an attempt to control existing knowledge of a video model, but an attempt to induce new knowledge into the model, which should require better and more diverse fine-tuning data.

\paragraph{Motion analysis limitations}
As discussed in \Cref{supp:motion-analysis-details}, we estimate the optical flows in the latent space of CogVideoX~\cite{CogVideoX} autoencoder rather than the pixel space, because it's the space the video DiT operates in and early trajectory steps produce disarranged decoder's outputs.
Besides (as also observed by \cite{GID}), motion spectral volumes are sensitive to the quality of optical flow estimation.
While the key conclusions (of the camera motion being low-frequency and kicking in very early in the diffusion trajectory) would hold since the are evident even with a bare eye from inspecting the denoising process visualization, the exact behavioral details of motion spectral volumes might change when an optical flow estimator is swapped or the analysis is moved from the latent to pixel space.

\paragraph{Linear probing limitations}
We inspect the presence of disentangled camera information in the video DiT model of a particular architecture and trained on particular data.
For it, we observe that the knowledge starts to arise from the $9$-th block, but for a different instance of a video model it can be distributed across the blocks differently.
Besides, we only evaluate it on RealEstate10K~\cite{RealEstate10k} test-set trajectories, and do not explore classical camera estimation datasets.
Such analysis was sufficient to draw actionable conclusions and improve our base method, but there is vast room in making it more rigorous.
We expect that video diffusion models are capable of revolutionizing the field of camera registration bringing strong priors about feature correspondences under difficult conditions, like scene motion, changing lighting and occlusions.

\section{CogVideoX Results}
\label{supp:cogvideox}

\begin{figure}
\centering
\includegraphics[width=0.8\linewidth]{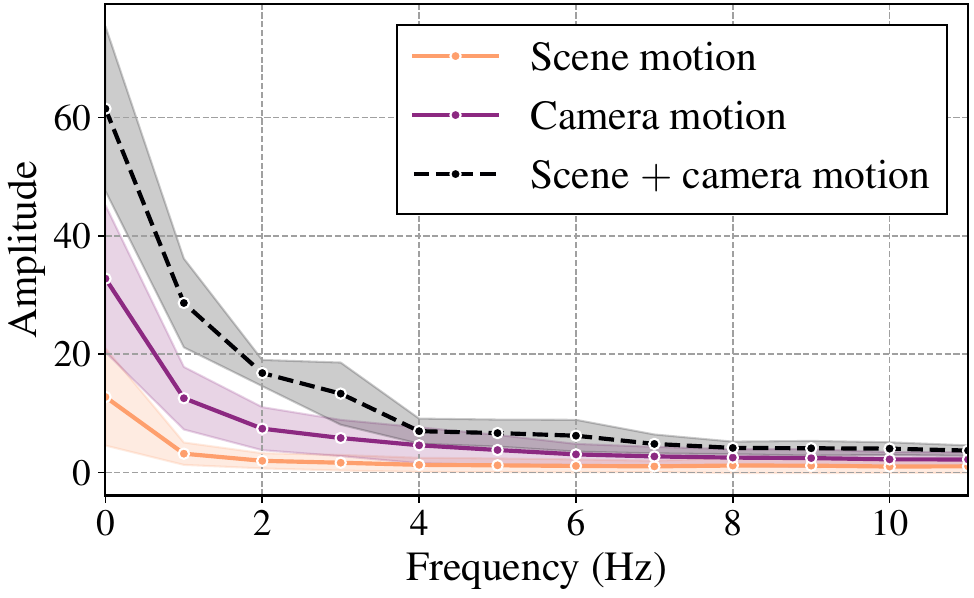}
\caption{
Comparing the average magnitude of motion spectral volumes for scenes with different motion types for CogVideoX~\cite{yang2024cogvideox}.
Videos with camera motion (purple) exhibit stronger overall motion than the videos with scene motion (orange), especially for the low-frequency range, suggesting that the motion induced by camera transitions is heavily biased towards low-frequency components.
}
\label{fig:spec-comparison-cogvideox}
\end{figure}

\begin{figure*}[t]
\centering
\begin{subfigure}[t]{0.32\linewidth}
\vspace{0.05cm}
\includegraphics[width=\linewidth]{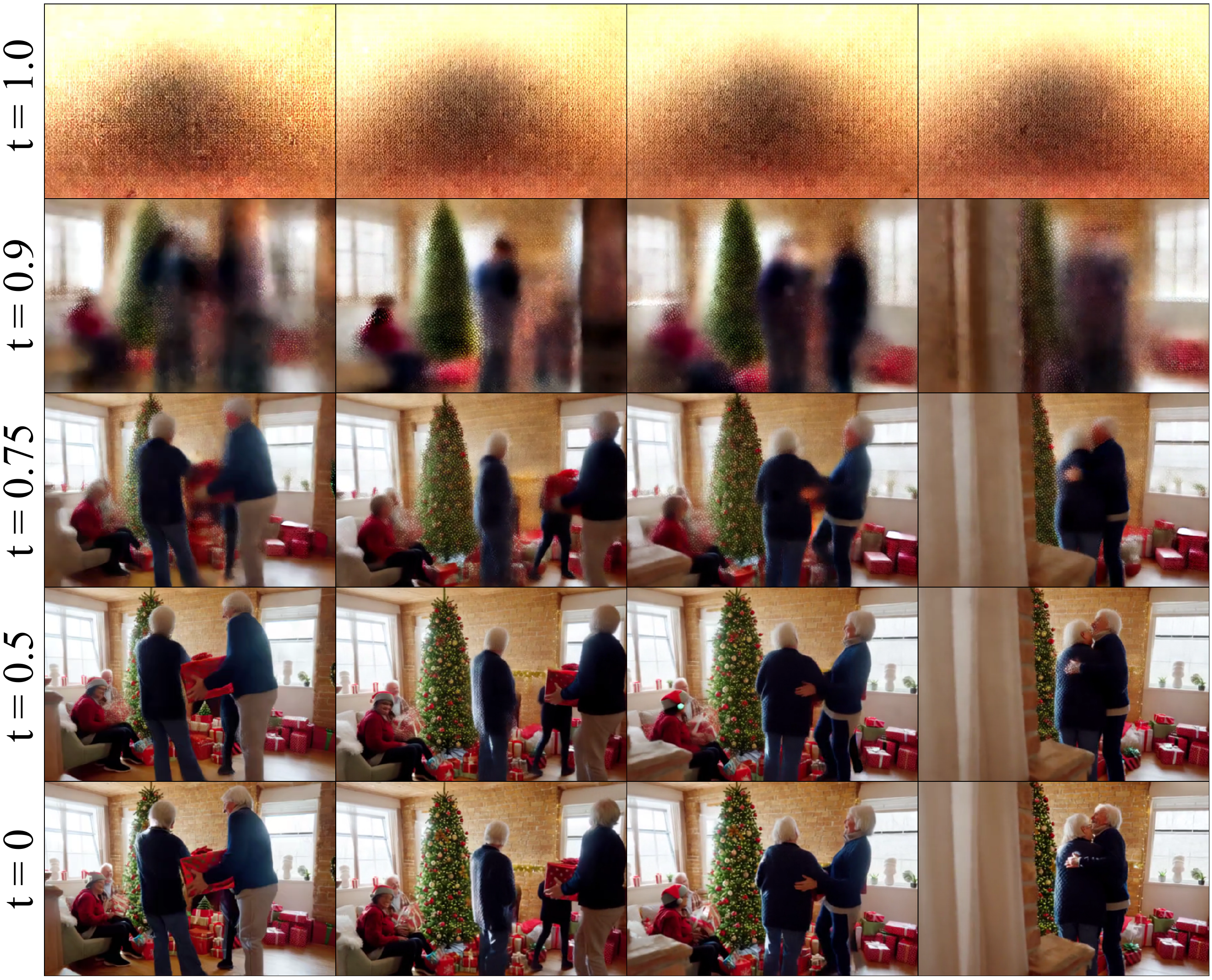}
\caption{A generated video at different diffusion timesteps. The camera has already been decided by the model even at $t = 0.9$ (first 10\% of the denoising process) and does not change after that.}
\label{fig:spec-analysis:visual-example-cogvideox}
\end{subfigure}%
\hfill
\begin{subfigure}[t]{0.66\linewidth}
\vspace{0pt}
\includegraphics[width=\linewidth]{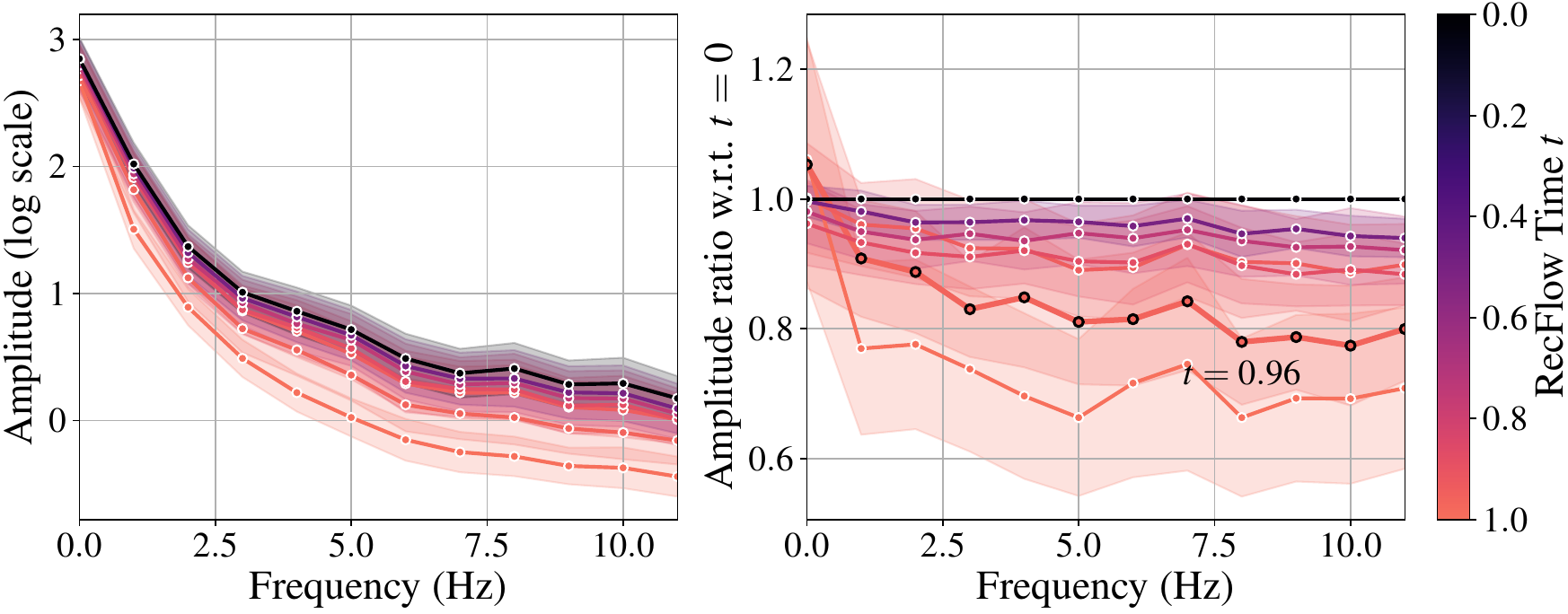}
\caption{Motion spectral volumes of \vditwithref's generated videos for different diffusion timesteps (left) and their ratio w.r.t. the motion spectral volume at $t=0$ (i.e., a fully denoised video).}
\label{fig:spec-analysis:spec-per-timestep-cogvideox}
\end{subfigure}%
\caption{
\textbf{How camera motion is modeled by diffusion (CogVideoX)?}
As visualized in \Cref{fig:spec-analysis:visual-example} and \Cref{fig:spec-comparison}, the motion induced by camera transitions is a low-frequency type of motion.
We observe that a video DiT creates low-frequency motion very early in the denoising trajectory: \Cref{fig:spec-analysis:spec-per-timestep} (left) shows that even at $t{=}0.96$ (first \apprx{4}\% of the steps), the low-frequency motion components have already been created, while high frequency ones do not fully unveil even till $t{=}0.5$.
We found that controlling the camera pose later in the denoising trajectory is not only unnecessary but \emph{detrimental} to both scene motion and overall visual quality. Here, we provide the same analysis conducted for CogVideox~\cite{yang2024cogvideox}.
}
\label{fig:spec-analysis-cogvideox}
\end{figure*}

We implement our method on top of CogVideoX~\cite{yang2024cogvideox} to show generalizability. Moreover, we conduct the motion analysis of the main paper for CogVideoX and show results in Fig.~\ref{fig:spec-comparison-cogvideox} and Fig.~\ref{fig:spec-analysis-cogvideox}. We observe a similar pattern, confirming the generalizability of our findings.

\section{Implementation details}
\label{supp:impl-details}

This section describes the training and architectural details of the base \vditwithref, \baselinewithref, and our downstream experiments.

\subsection{\vdit implementation details}

\paragraph{Architecture details}
Our video DiT~\cite{DiT} architecture follows a very similar design to the other contemporary video DiT models (e.g., \cite{Sora, MovieGen, OpenSora, LuminaT2X, CogVideoX}.
As the backbone, we used a transformer-based architecture of 32 DiT blocks~\cite{DiT}.
Each DiT block consists on a cross-attention layer to read the text embeddings information, produced by the T5~\cite{T5} model; a self-attention layer, and a fully-connected network with a $4 \times$ dimensionality expansion.
Each attention layer has 32 heads and RMSNorm~\cite{RMSNorm} for queries and keys normalization.
To encode positional information, we used 3D RoPE~\cite{RoPE} attention, where each axis (temporal, vertical, and horizontal) had a fixed dimensionality allocated for it in each attention head (we split the dimensions in the ratio of 2:1:1 for temporal, vertical, and horizontal axes, respectively).
LayerNorm~\cite{LayerNorm} is used to normalize the activations in each DiT block.
We used CogVideoX~\cite{CogVideoX} autoencoder which is a causal 3D convolutional autoencoder with $4\times8\times8$ compression rate and 16 channels for each latent token.
The hidden dimensionality of our DiT model is 4,096 and it has \paramcount\ parameters in total.
Similar to DiT~\cite{DiT}, we use block modulations to condition the video backbone on the rectified flow timestep information, SiLU~\cite{GELU} activations and $2 \times 2$ ViT-like~\cite{ViT} patchification of the input latents to reduce the sequence size.

\paragraph{Training details}
We train the model with the AdamW~\cite{AdamW} optimizer with the learning rate of 0.0001 and weight decay of 0.01.
The model was trained for 750,000 total iterations with the cosine learning rate scheduler~\cite{SGDR} in \textsf{bfloat16}.
We also incorporate the support of image animation by encoding the first frame with the same CogVideoX encoder, adding random gaussian noise (with the noise level sampled independently from the video noise levels $\sigma_t$), projecting with a separate learnable ViT-like~\cite{ViT} patchification layer, repeating sequence-wise to match the video length and summing with the video tokens.
During training, we use loss normalization~\cite{EDM2}.
The model is trained jointly on images and videos of variable resolution ($256$, $512$ and $1024$), aspect ratio ($16:9$ and $9:16$ for videos, and $16:9$, $9:16$ and $1:1$ for images), and video lengths (from $17$ frames to $385$ frames).
The video framerate was set to 24 frames per second and we did not use variable-FPS training as contemporary works~\cite{SnapVideo,MovieGen} since we found it to decrease the performance for a target framerate (at least, without fine-tuning).

\paragraph{Inference details}
During inference, we use the standard rectified flow without any stochasticity.
We found that 40 steps gives a good trade-off between quality and sampling speed.
We follows the same time shifting strategy as Lumina-T2X for higher resolutions and longer video generation~\cite{LuminaT2X} with time shifting of $\sqrt{32}$ for the $1024$ resolution.

\subsection{\baselinename implementation details}

As being said in \Cref{sec:method:baseline-cc}, \baselinename is a simple ControlNet-like~\cite{ControlNet, ControlNet-pp} fine-tuning of \vdit for camera control.
We use smaller versions of the base \vdit blocks with only a 128 hidden dimensionality and 4 attention heads.
Besides, we do not use cross-attention over the context information since we found it to severely decrease the visual quality and camera control precision.
For the \Plucker encoding computation, we replicate the pipeline of VD3D~\cite{VD3D}.
Our linear layer that processes them contains 4096 hidden features and SiLU~\cite{GELU} non-linearity.

\subsection{\methodname implementation details}

We train our complete setup for 6K iterations on the joint dataset of 65K videos from \retenkfull~\cite{RealEstate10k} and 20K dynamic videos with static cameras (this dataset is described in \Cref{sec:method:data} and \Cref{supp:data-details}).
We train with the learning rate of 0.0001 using the AdamW~\cite{AdamW} optimizer with weight decay of 0.01 and cosine learning schedule~\cite{SGDR}.

\begin{figure}
\centering
\includegraphics[width=3in]{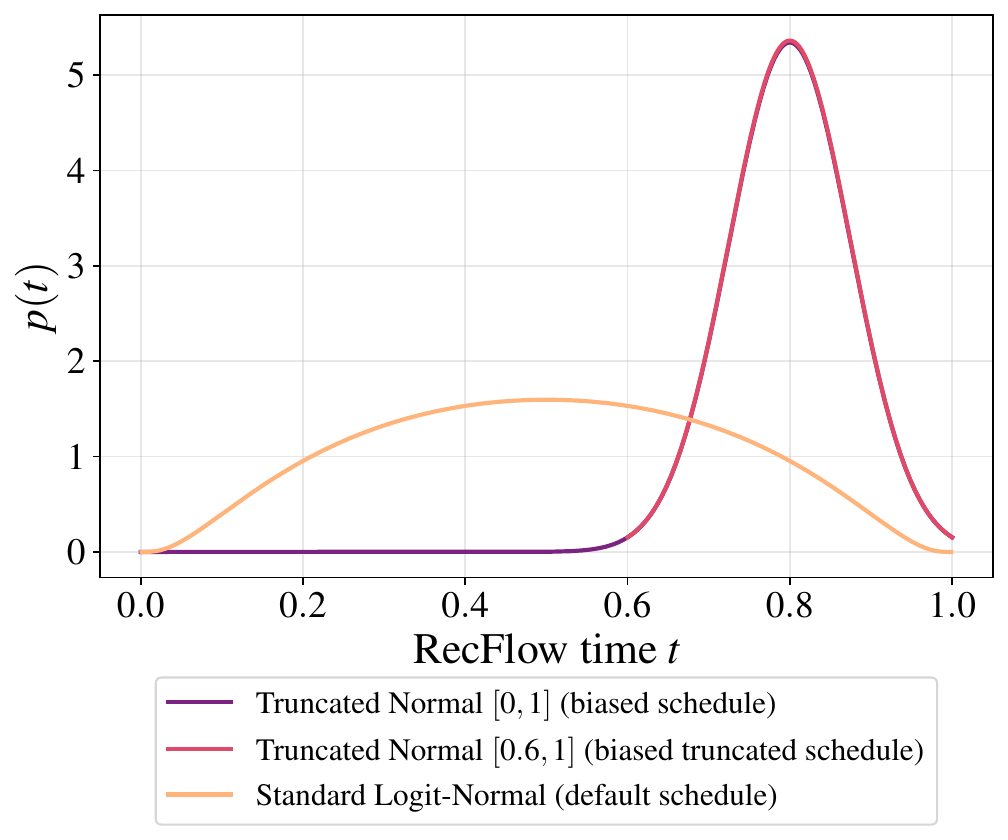}
\caption{Comparing rectified flow noise schedules: (orange) vanilla standard logit-normal noise schedule proposed by \cite{SD3} and used for baseline experiments; (purple) biased but non-truncated noise schedule; (pink) biased and truncated noise schedule.}
\label{fig:noise-schedules}
\end{figure}

As described in \Cref{sec:method:baseline-cc}, since the camera motion is a low-frequency type of signal, we propose to use truncated and biased noise schedule: both at train and inference time.
In \Cref{fig:noise-schedules}, we visualize three distributions: 1) the standard one used by SD3~\cite{SD3}, our base \vdit, and ``w/o biasing noise''  experiment (orange); 2) the biased but non-truncated schedule used in the \baselinename ``w/o noise truncation'' ablation experiment (purple), which has some unnecessary and detrimental probability mass in the high-frequency range; and 3) our final schedule (red).

Each ablation experiment from \Cref{sec:experiments:ablations} was trained for 6K iterations on 32 NVIDIA A100 80GB GPUs.

During training, to plug in the conditioning camera pose for our static dataset, we were randomly sampling extrinsics and intrinsics parameters from a \retenkfull dataset.

\paragraph{1D temporal camera encoder}
\baselinewithref processes videos in the autoencoder's latent space with $4\times$ temporal compression~\cite{CogVideoX}, raising the question of how to incorporate full-resolution camera parameters.
While camera poses could be naively downsized (e.g., subsampled or reshaped) to match the latent resolution, this would force small \ditxs\ blocks to process compressed camera information.
Instead, we implement a sequence of causal 1D convolutions that transform a $F {\times} 6$ sequence of \Plucker\ coordinates for each pixel into a $(F // 4) {\times} 32$ representation.

\section{Motion analysis details}
\label{supp:motion-analysis-details}

As discussed in \Cref{sec:method:motion}, we perform camera motion analysis of generated videos at different time steps by inspecting their spectral volumes.

\begin{figure}
\centering
\includegraphics[width=\textwidth]{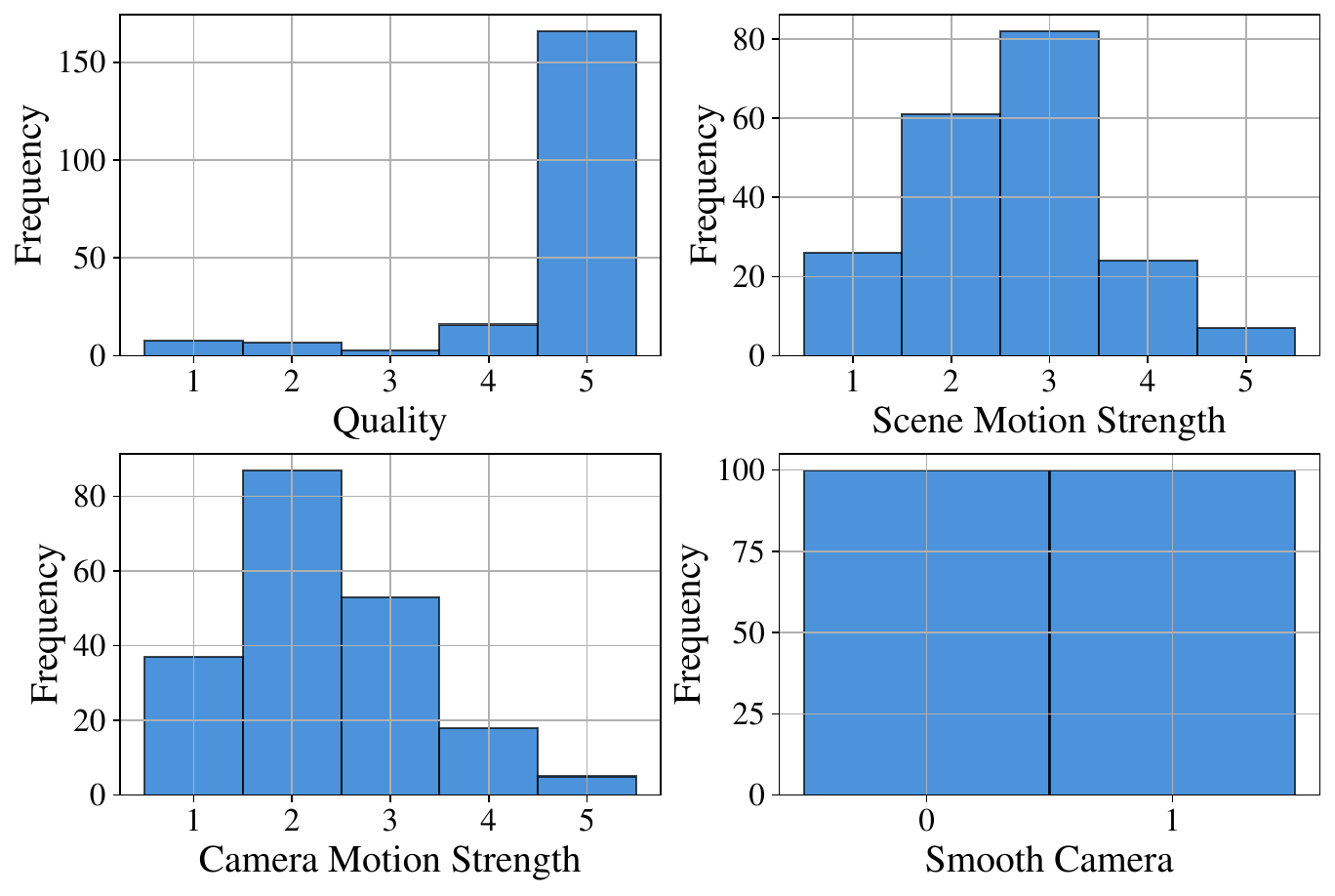}
\caption{Our annotations collected for 200 randomly generated videos from \vditwithref and used in our camera motion analysis in \Cref{sec:method:motion}.}
\label{fig:camera-motion-annotation}
\end{figure}

For our analysis, we generate 200 random 121-frames videos without time shifting~\cite{LuminaT2X} in the $288\times 512$ resolution with \vdit and manually annotate them for the following labels: quality (a score from 1 to 5), scene motion strength (a score from 1 to 5, where a score of 1 corresponds to a completely still scene), camera motion strength (a score from 1 to 5, where a score of 1 corresponds to a completely stationary viewpoint), whether the camera is smooth or shaky (a binary flag).
We visualize the obtained scores in \cref{fig:camera-motion-annotation}.
Next we discard the videos with the quality score of less than 4, discarding 18 out of 200 videos: ``broken'' samples (e.g., an artificial animation or a blank black canvas) indicate a complete generation failure which we should exclude from the analysis of the camera motion.

To analyze the spectral volumes differences between scene and camera motion for \cref{fig:spec-comparison}, we extract three categories of videos from our dataset: 1) scene motion (videos with scene motion, but with no camera motion); 2) camera motion (videos with camera motion, but with no scene motion); and 3) scene and camera motion (videos with scene motion and non-artifactory camera motion).
The first category (scene motion) was obtained by selecting the videos with the camera motion strength of 1 (i.e., no camera movement), and the scene motion strength of more or equal than 3; the second category (camera motion) was obtained by selecting the videos with the camera motion strength of more or equal than 3, and the scene motion strength of 1 (i.e., no scene motion); the third category (scene and camera motion) was obtained by selecting the videos with the camera and scene motion strengths of more or equal than 3 and the smooth camera flag being true (to exclude shaky camera movements).
To analyze the spectral volumes for \cref{fig:spec-analysis}, we took the videos which have scene or camera motion strength of more or equal than 3.

\begin{figure}
\centering
\includegraphics[width=\textwidth]{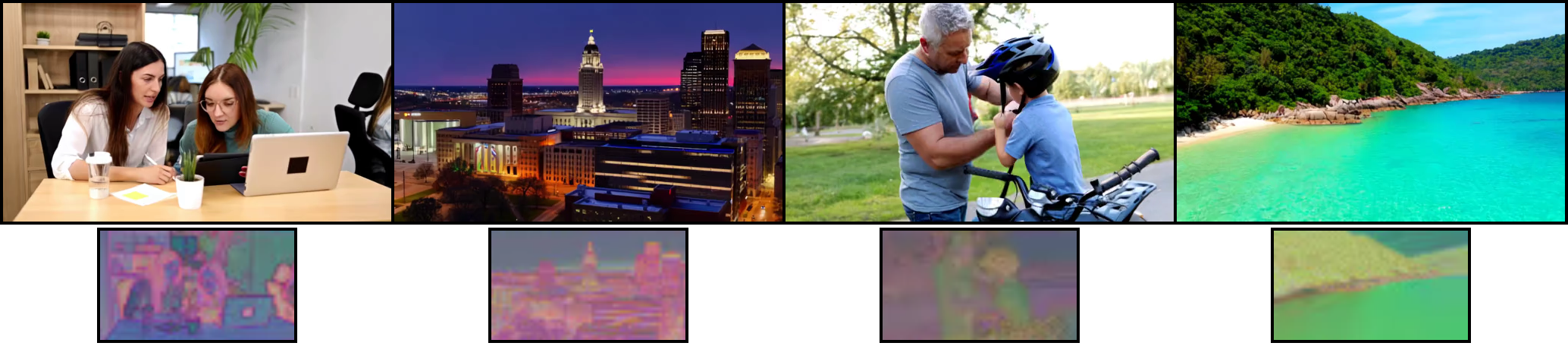}
\caption{Frames of the generated videos by the \vditwithref model (upper row) and the corresponding PCA projections of their latents (lower row).}
\label{fig:latents-pca}
\end{figure}

To obtain spectral volumes, we need to obtain per-pixel optical flow information.
Since our \vditwithref is following the latent diffusion (LDM) paradigm~\cite{LatentDiffusion}, we opt for performing optical flow estimation in the latent space of the autoencoder.
There are two reasons for that: 1) we noticed that it provides more robust flow estimation at earlier denoising timesteps (since the decoder part of CogVideoX~\cite{CogVideoX} autoencoder does not need to operate at out-of-distribution inputs); and 2) the video model operates in this space.
In this way, we used the raw generated latents to estimate the optical flow.
Since most of the optical flow algorithms operate in a 3-dimensional RGB space, and our latents are 16-channels, we projected them into 3-dimensional inputs via a PCA, computing it independently for each latent.
We found that these representations maintain very strong spatiotemporal resemblance to the original videos, as visualized in \cref{fig:latents-pca}.

Following \cite{GID}, we use PyFlow~\cite{PyFlow} coarse-to-fine optical flow estimation to obtain more robust results.
We attempted to use Farneback~\cite{FarnebackOF} optical flow estimation, but observed that it is less accurate and does align less with our visual evaluation of the results.
121-frames $288 \times 512$-resolution videos correspond to 31 latent frames of $36 \times 64$ resolution.
For each 6-th latent frame, we estimate its flow with respect to each of the 24 subsequent frames.
Next, we perform Fast Fourier transform for $x$ and $y$ spatial coordinates independently, compute the amplitudes and average them spatially for each video.

\section{Linear probing details}
\label{supp:linprob-details}

In \Cref{sec:method:linprob}, we perform linear probing of \vditwithref for camera pose knowledge.
For this, we use 1,000 videos of $49$ frames and the $144 \times 256$ resolution from the test set of \retenkfull~\cite{RealEstate10k} and extract their internal representations of our \vdit model under various noise levels.
We use the noise levels $\sigma_t$ of $[\frac{1}{8}, \frac{2}{8}, \frac{3}{8}, ..., \frac{7}{8}, 1]$.
The hidden dimensionality of our \vdit model is $4,096$ and to reduce the memory requirements and speed up linear probing we project each video representations into 512 dimensionality using PCA.
This results in latent representations of $512 \times 13 \times 18 \times 32$ for each block, each video and each timestep.
To construct the training features for the linear regression model, we extract the (spatially) middle vector of shape $512 \times 13$ and perform spatial average pooling to obtain a context representation of $512 \times 13$.
We then unroll and concatenate them to obtain the final training representation of dimensionality $13,312$.
We split our 1,000 videos into training and test sets as 900 and 100 and then train a ridge linear regression with the regularization weight of 25,000.
Our target variable covers the extrinsics parameters of the viewpoints, which are provided by \retenk.
We extract rotation angles and translations from the extrinsic matrices and normalize them with respect to the first frame.
Then, we compute the rotation and translation errors on the held-out set of 100 videos using the evaluation pipeline of CameraCtrl~\cite{CameraCtrl}.
Since we have 8 noise values and 32 \vdit blocks, this resulted into 256 linear regression models in total.
It was taking \apprx{5} minutes of CPU time to train each model, and their training was parallelizable across different cores.

\section{Dataset construction details}
\label{supp:data-details}

As describe in \Cref{sec:method:data}, we construct a dataset of videos with scene motion but recorded from stationary cameras.
One might attempt to build such a dataset in an automated way by estimating the optical flow and checking whether there is non-zero motion in the middle region of the frame, and no motion on the borders.
However, this would not work well for a myriad of corner cases, which is why we opted for manual construction.
For this, we annotated 110K internal videos for the presence of camera and the presence of scene motion.
Each video has a duration of $5-30$ seconds and covers various diverse categories of scenes: humans, animals, landscapes, food and other types.
Out of these videos, 20K videos turned out to be the necessary ones: with scene motion, but completely static scenes.
We proceeded to use them as our training data without further processing.
As being discussed in \Cref{supp:impl-details}, we augmented their camera information artifically by using random extrinsics from \retenkfull~\cite{RealEstate10k}.

\section{Scaling bias}
\label{supp:scaling-bias}
Classical feature-based camera estimation~\cite{colmap_sfm, colmap_mvs} outputs camera trajectories with arbitrary scale, as it does not possess any priors about the absolute size of objects in the scene. For example, a house might appear identical to a small object like an apple when viewed from the appropriate distance, making it impossible to determine absolute scale from visual features alone. This lack of scale awareness complicates precise user control over camera trajectories: while the model can determine the camera's direction, it remains unaware of the magnitude of each movement, as demonstrated in~\cref{sec:experiments}.
Ideally, all camera trajectories should be aligned to a consistent reference scale.
A natural reference would be a metric scale, now achievable due to recent advances in metric depth estimation \cite{Metric3D}.
We propose a method to rescale camera trajectories across all data using the following steps:
\begin{enumerate}
\item Obtain 3D points from the COLMAP output and render the COLMAP depth $D^f_c$ from these points for each frame.
\item Estimate the metric depth $D^f_m$ for each frame using a pre-trained zero-shot metric depth estimator~\cite{Metric3D}.
\item Calculate the re-scaling parameter $\hat{\lambda}$ by solving the optimization problem $\hat{\lambda} = \arg\min_\lambda \mathbb{E}_{f \sim F} \left| \lambda D^f_c - D^f_m \right|$, where $F$ is the total number of frames.
\item Set the camera translation vector $\hat{\bm{t}}_c$ to $\hat{\lambda} \bm{t}_c$.
\end{enumerate}
In \cref{sec:experiments:ablations}, we show that training with properly scaled cameras does not lead to visual quality degradation (even improving it slightly), and, as we demonstrate in our supplementary visuals, makes the camera control more predictable and less frustrating for a user by allowing to control the magnitudes of camera transitions.

\section{Additional Related Work}

Due to space constraints, we summarize related 3D and an extended list of 4D works in the appendix.

\inlinesection{3D generation.} Early efforts in 3D generation focused on training models for single object categories, extending GANs to 3D by incorporating neural renderers as an inductive bias~\cite{devries2021unconstrained, chan2022efficient, or2022stylesdf, schwarz2022voxgraf, bahmani2023cc3d}. As the field progressed, CLIP-based supervision~\cite{radford2021learning} enabled more flexible and diverse 3D asset generation, supporting both text-based generation and editing~\cite{chen2018text2shape, jain2022zero, sanghi2022clip, jetchev2021clipmatrix, gao2023textdeformer, wang2022clip}. Recent advances in diffusion models further enhance generation quality by replacing CLIP with Score Distillation Sampling (SDS) for supervision~\cite{poole2022dreamfusion, wang2023prolificdreamer, lin2022magic3d, chen2023fantasia3d, liang2023luciddreamer, wang2023score, li2024controllable, he2024gvgen, ye2024dreamreward, liu2023humangaussian, yu2023text, katzir2023noise, lee2023dreamflow, sun2023dreamcraft3d}.
To improve the structural coherence of 3D scenes, several approaches generate multiple views of a scene for consistency~\cite{lin2023consistent123, liu2023zero, shi2023mvdream, feng2024fdgaussian, liu2024isotropic3d, kim2023neuralfield, voleti2024sv3d, hollein2024viewdiff,tang2024pixel,gao2024cat3d}. Alternatively, iterative inpainting has been explored as a technique for scene generation~\cite{hollein2023text2room, shriram2024realmdreamer}.
Recent works also focus on lifting 2D images to 3D representations, employing methods like NeRF~\cite{NeRF}, 3D Gaussian Splatting~\cite{kerbl20233d}, or meshes in combination with diffusion models~\cite{chan2023generative, tang2023make, gu2023nerfdiff, liu2023syncdreamer, yoo2023dreamsparse, tewari2024diffusion, qian2023magic123, long2023wonder3d, wan2023cad, szymanowicz2023viewset}. Other studies explore fast, feed-forward 3D generation techniques that directly predict 3D models from input images or text~\cite{hong2023lrm, li2023instant3d, xu2023dmv3d, xu2024grm, zhang2024compress3d, han2024vfusion3d, jiang2024brightdreamer, xie2024latte3d, tang2024lgm, tochilkin2024triposr, qian2024atom, szymanowicz2023splatter, szymanowicz2024flash3d}. These methods, however, are limited to synthesizing static scenes, in contrast to our approach.

\inlinesection{4D generation.} There has been significant progress in 4D generation, i.e., dynamic 3D scene generation. These works often rely on input text prompts or images to guide the generation. Since the early advancements in large-scale generative models for this task~\cite{singer2023text}, significant strides have been made in improving both the visual and motion quality of generated scenes~\cite{ren2023dreamgaussian4d, ling2023align, bahmani20234d, zheng2023unified, gao2024gaussianflow, yang2024beyond, jiang2023consistent4d, miao2024pla4d, li2024vivid,zhang4900422motion4d,yuan20244dynamic,jiang2024animate3d}.
While many of these methods are conditioned on text input, other approaches focus on converting 2D images or videos into dynamic 3D scenes~\cite{ren2023dreamgaussian4d, zhao2023animate124, yin20234dgen, pan2024fast, zheng2023unified, ling2023align, gao2024gaussianflow, zeng2024stag4d, chu2024dreamscene4d, wu2024sc4d, yang2024diffusion, wang2024vidu4d, feng2024elastogen, sun2024eg4d, zhang20244diffusion, ren2024l4gm, lee2024vividdream, li20244k4dgen, van2024generative, uzolas2024motiondreamer, chai2024star, liang2024diffusion4d, li2024dreammesh4d,SV4D}.
Recently, several works~\cite{lin2024phy124,huang2024dreamphysics,zhang2024physdreamer} investigate physics priors in 4D generation pipelines.
Other works~\cite{zhang2024magicpose4d,sun2025ponymation,chen2024ct4d} enhance motion controllability with template-based methods.
Another line of work~\cite{bahmani2024tc4d,xu2024comp4d,cao2024avatargo,yu20244real,zeng2024trans4d,zhu2024compositional} focuses on compositional and interactive 4D generation.
Another strand of research extends 3D GANs into the 4D domain by training on 2D video data~\cite{bahmani20223d, xu2022pv3d}. However, the quality of these methods is often constrained by the limited nature of the datasets, which typically focus on single object categories. Moreover, the majority of these approaches tackle object-centric generation. As a result, they typically neglect background elements, and their visual fidelity falls short when compared to the high photorealism achieved by state-of-the-art video generation models, such as those employed in our approach.

\inlinesection{Motion-controlled video generation.}
Orthogonal to camera-controlled video generation methods, another line of work investigates object trajectory control~\citep{wang2023videocomposer,yin2023dragnuwa,wu2024draganything,zhou2024trackgo,wang2024boximator,zhang2024tora}.
More recently, several methods~\citep{namekata2024sg,qiu2024freetraj,ma2023trailblazer,DirectAVideo,jain2024peekaboo,Img2VidAnim-Zero} focus on object trajectory control without relying on additional external data or additional model fine-tuning.
\section{User Study}
\label{supp:user_study}

In the user study, we engage 10 professional labelers, each of whom evaluates 100 different video pairs.
The labelers are asked to choose between two videos based on several preference metrics:
\begin{itemize}
    \item \textit{Camera Alignment (CA):} How well the camera trajectory follows the reference video.
    \item \textit{Motion Quality (MQ):} Which video has larger and more natural motion.
    \item \textit{Text Alignment (TA):} Which video better aligns with the provided reference text prompt.
    \item \textit{Visual Quality (VQ):} Which video has a higher overall visual quality.
    \item \textit{Overall Preference (Overall):} Which generated video the user would prefer for this task.
\end{itemize}


\end{document}